\documentclass[review]{elsarticle}

\usepackage{hyperref}

\usepackage{amssymb}
\usepackage{enumerate}
\usepackage{graphicx}%
\usepackage{hyperref}
\hypersetup{hypertex=true,
	colorlinks=true,
	linkcolor=blue,
	anchorcolor=blue,
	citecolor=blue}
\usepackage{multirow}%
\usepackage{array}%
\usepackage{diagbox}%
\usepackage{epstopdf}%
\usepackage{booktabs}%
\usepackage{manyfoot}%
\usepackage{listings}%
\usepackage{makecell}%
\usepackage{longtable}%
\usepackage{threeparttable}%
\usepackage{amsmath,amssymb,amsfonts}%
\usepackage{mathrsfs}%
\usepackage{xcolor}%
\usepackage{textcomp}%
\usepackage{algorithm}%
\usepackage{algorithmicx}%
\usepackage{algpseudocode}%
\usepackage{float}%
\usepackage{bm}%
\usepackage[section]{placeins}

\biboptions{authoryear}

\begin{document}

\begin{frontmatter}

\title{Automated Novelty Evaluation of Academic Paper: A Collaborative Approach Integrating Human Expertise and Large Language Models}
\author{Wenqing Wu}
\ead{winchywwq@njust.edu.cn}

\author{Chengzhi Zhang \corref{cor1}}
\ead{zhangcz@njust.edu.cn}

\author{Yi Zhao}
\ead{yizhao93@njust.edu.cn}

\cortext[cor1]{Corresponding author}
\affiliation{organization={Department of Information Management, Nanjing University of Science and Technology},
	city={Nanjing},
	postcode={210094}, 
	country={China}}

\begin{abstract}
Novelty is a crucial criterion in the peer review process for evaluating academic papers. Traditionally, it's judged by experts or measure by unique reference combinations. Both methods have limitations: experts have limited knowledge, and the effectiveness of the combination method is uncertain. Moreover, it's unclear if unique citations truly measure novelty. The large language model (LLM) possesses a wealth of knowledge, while human experts possess judgment abilities that the LLM does not possess. Therefore, our research integrates the knowledge and abilities of LLM and human experts to address the limitations of novelty assessment. One of the most common types of novelty in academic papers is the introduction of new methods. In this paper, we propose leveraging human knowledge and LLM to assist pretrained language models (PLMs, e.g. BERT etc.) in predicting the method novelty of papers. Specifically, we extract sentences related to the novelty of the academic paper from peer review reports and use LLM to summarize the methodology section of the academic paper, which are then used to fine-tune PLMs. In addition, we have designed a text-guided fusion module with novel Sparse-Attention to better integrate human and LLM knowledge. We compared the method we proposed with a large number of baselines. Extensive experiments demonstrate that our method achieves superior performance.
\end{abstract}


\end{frontmatter}

\section{Introduction}
\label{intro}
Novelty is one of the most crucial criteria in academic publishing. In an academic context, novelty is defined as the reorganization of existing knowledge in an unprecedented manner \citep{r1,r2}. Based on this definition, researchers consider assessing the novelty of the academic paper to involve the identification of knowledge reorganization. They recognize that knowledge reorganization is rooted in the content of the reference section and can be measured by novel combinations of references or journal pairs in the reference list to gauge the research's novelty \citep{r3,r4,r5}. These methods have certain drawbacks. Firstly, the extent to which cited publications serve as sources of inspiration for a paper is not yet well understood \citep{r6}. Secondly, there is citation bias, as some fields or disciplines tend to cite classic or traditional literature, overlooking more recent research. Additionally, citations may serve purposes such as criticism, rebuttal, or analysis. Therefore, it is essential to make timely assessments of the novelty of research without relying solely on citations. With the advancement of the open-access movement, access to full-text academic papers, peer reviews, and related materials has become more accessible. These textual resources contain a wealth of information, offering new data sources and evaluation dimensions for assessing academic papers. Because peer review \citep{r7} necessitates a comprehensive evaluation of a manuscript's novelty, reliability, clarity, and significance to maintain consistently high standards in academic publications. The rapid advancement of artificial intelligence \citep{r8,r9,r10} has also opened possibilities for predicting the novelty of papers without relying on citations.\\
\begin{figure*}[h]%
	\centering
	\includegraphics[width=1\textwidth]{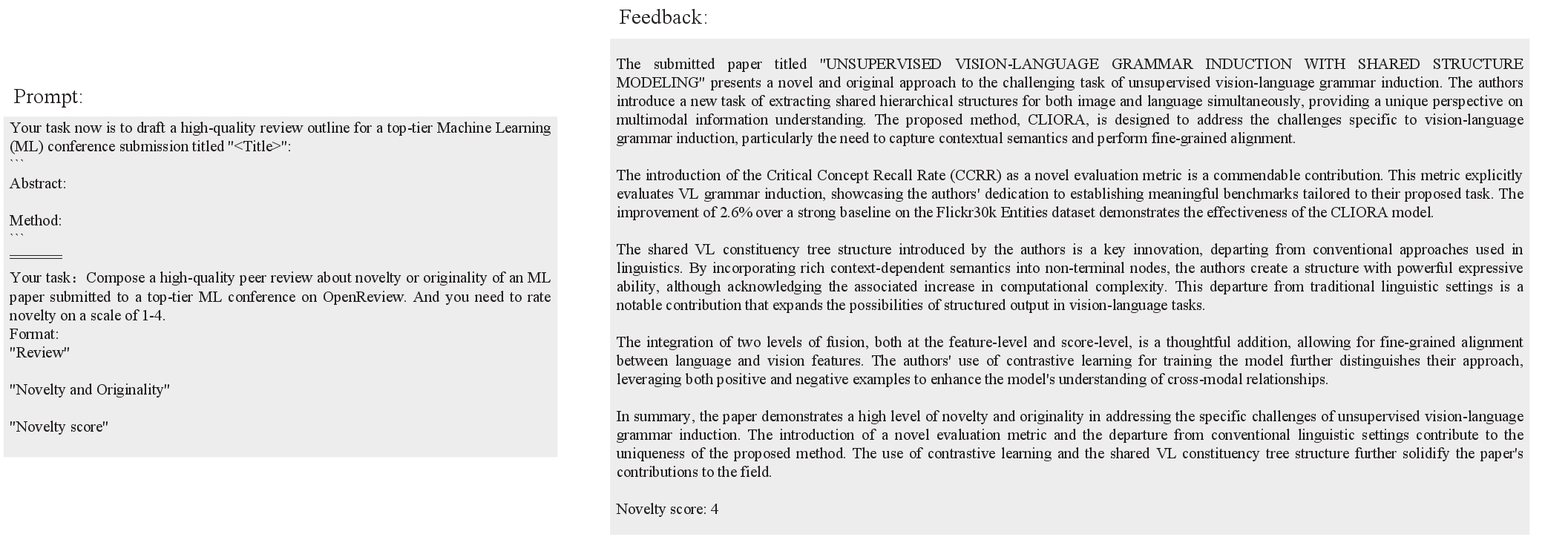}
	\vspace{-0.7cm}
	\caption{\centering{ChatGPT provides feedback and scores for the novelty of the paper's methodology. These paper’s novelty score is 3. On the left are the prompts, and on the right are the feedback responses.}}
	\label{fig:1}
\end{figure*}

\indent In the wave of enthusiasm for large language models (LLMs) \citep{r11,r12}, researchers have been actively exploring various application areas \citep{r13,r14,r15,r16}, such as automating paper screening \citep{r17}, error identification \citep{r18}, and checklist verification \citep{r19}. Meanwhile, Liang et al. \citep{r20} conducted a large-scale empirical analysis on whether LLMs can provide useful feedback for research papers. Mike \citep{ref61} believes that ChatGPT still seems to lack accuracy and is not enough for any formal or informal research quality assessment tasks. Based on their research findings, LLMs face challenges in independently assessing the novelty of papers. On this basis, we conducted tests on LLMs regarding the novelty evaluation of papers. We use papers with various expert-assigned novelty scores as input. As depicted in Figure \ref{fig:1}, we found that LLMs possess summarization capabilities but lack the ability to assess novelty. Because regardless of whether an expert-assigned novelty score for a paper is high or low, the LLMs always assigns it the highest novelty score. Feedback for other scores is presented in Figure \ref{fig:s1}, \ref{fig:s2} and \ref{fig:s3} in the Appendix \ref{app1}. While human expert reviewers generally possess the ability to assess the novelty of papers and often articulate their evaluations in the review reports. Hence, how to effectively harness the capabilities of human reviewers and LLMs to assist deep learning models in obtaining real-time assessment on novelty is a question worthy of research. Previous researchs \citep{r43,r44} on novelty prediction has predominantly focused on the degree of novelty, without distinguishing between the types of novelty.\\
\begin{figure*}[h]%
	\centering
	\includegraphics[width=0.7\textwidth]{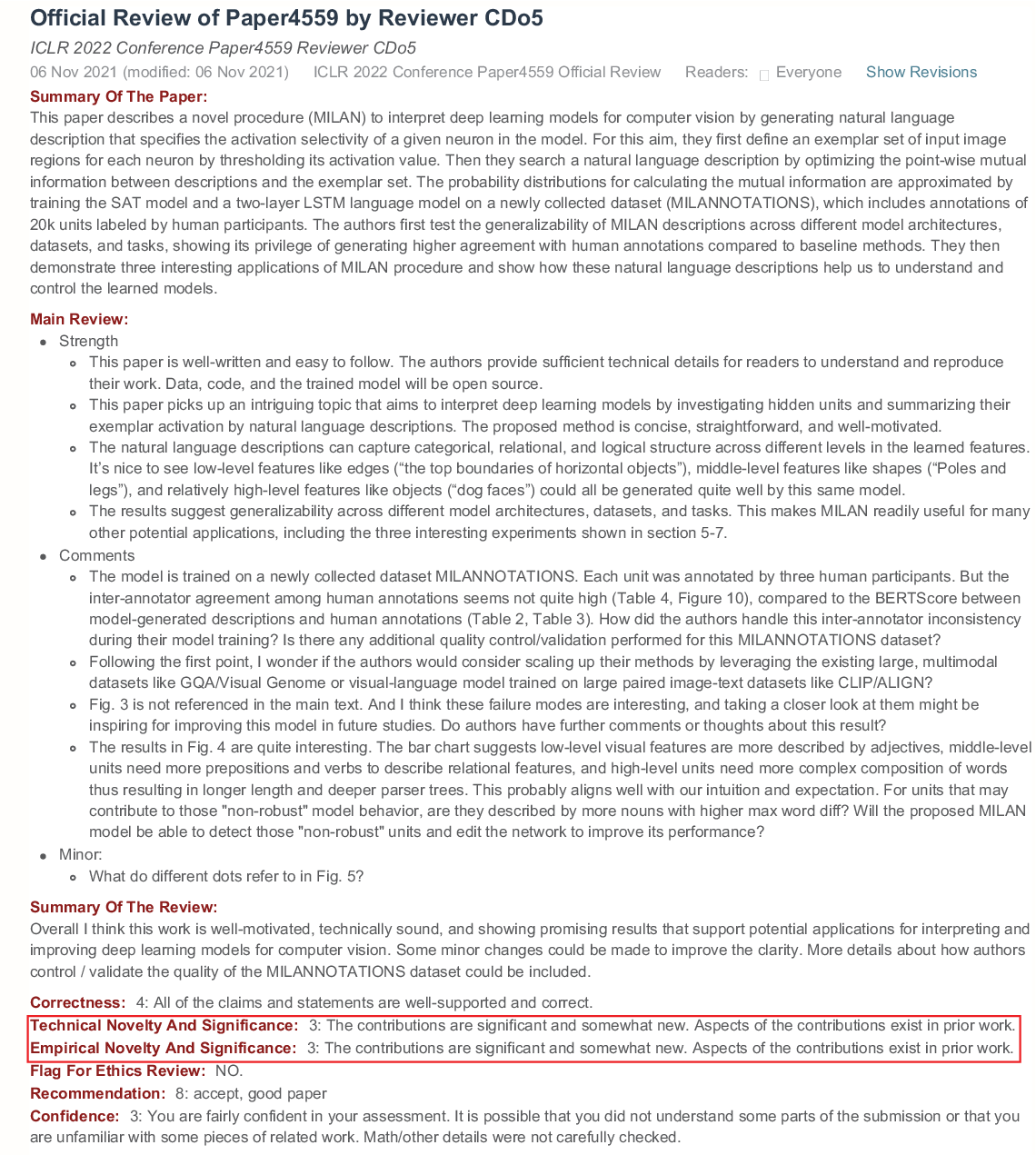}
	\vspace{-0.1cm}
	\caption{\centering{An Example of review report on ICLR 2022. (\href{https://openreview.net/forum?id=NudBMY-tzDr}{https://openreview.net/forum?id=NudBMY-tzDr})}}
	\label{fig:2}
\end{figure*}

\indent To delve into the aforementioned question, we need to obtain data related to novelty. Currently, there is no unified gold standard dataset for comprehensive evaluation of novelty, whether it pertains to theoretical novelty, method novelty, or result novelty. Recently, Leahey et al. \citep{r62} noted in their study on types of novelty in academic papers that one of the most common type of novelty is the introduction of new methods, which appears in 57\% of the papers analyzed. Starting from ICLR 2022, in addition to soliciting recommendations and confidence score from reviewers, the requirement for reviewers to provide scores on Technical/Empirical Novelty and Significance has also been introduced, as shown in Figure \ref{fig:2}. Considering that ICLR is a top-tier conference in the field of deep learning, where research typically focuses on the model level (most of which pertains to method novelty), and given that Technical Novelty and Significance (TNS) is defined as new models, techniques, or theoretical insights\footnote{\href{https://www.cs.virginia.edu/~hw5x/Course/RL2022-Fall/_site/readings/}{https://www.cs.virginia.edu/~hw5x/Course/RL2022-Fall/\_site/readings/}}, we believe TNS can serve as a standard for evaluating method novelty. Subsequently, we extract the methodology sections from academic papers and input them into ChatGPT\footnote{\href{https://chat.openai.com/}{https://chat.openai.com/}} to obtain summaries regarding the novelty of the methodologies. Because the gold standard we are studying is Technical Novelty and Significance score, we believe that most of the technical novelty focus in the ICLR conference should be on the methodology section, so we only used the methodology section as input. Simultaneously, we extract sentences related to novelty from peer review reports as a human assessment of novelty. We input both sets of texts into a pre-trained language model to predict the novelty of the methodologies in academic papers. Additionally, we use the original text of the methodology sections from academic papers and the sentences indicating novelty from review reports as training corpora for prediction. The results indicate that models trained on text summarized by LLMs exhibit superior performance. In order to better leverage human and LLMs knowledge, we have designed a text-guided fusion module with novel sparse attention to integrate the knowledge of both. The experimental results indicate that with the assistance of this module, the model significantly improved in performance. \\
\indent In summary, the contributions of this paper include four aspects:\\
\indent This paper proposed a novel approach for predicting the method novelty of academic papers.\\
\indent This paper explored a novel method to leverage the knowledge of both human and LLMs in aiding deep learning models to achieve enhanced performance.\\
\indent This paper designed a text-guided fusion module to guide and integrate the knowledge of both human and LLMs, enabling effective utilization of the knowledge.\\
\indent The thorough experiments show that our method is effective, and has achieved commendable results. The code and dataset for this paper can be accessed at \href{https://github.com/njust-winchy/method_novelty_predict}{https://github.com/njust-winchy/method\_novelty\_predict}.

\section{Related Work}
\label{related}
In this section, we will report related work about our research. Firstly, we present relevant work in the field of automated scholarly peer review. Subsequently, we delve into literature concerning novelty evaluation, followed by an overview of studies investigating the application of LLMs in the context of peer review.

\subsection{Automated Reviewing of Academic Paper}
Peer review is the cornerstone of science and plays a crucial role in the scientific communication system \citep{r21}. Research on peer review has long been an intriguing topic that has captured the attention of scholars \citep{r22,r23,r24,r25,r26}. With the advancement of natural language processing (NLP) technologies, an increasing number of researchers have delved into the realm of automated reviewing of academic paper \citep{r27}, addressing topics such as predicting decisions on academic paper submissions \citep{r28} and mining arguments \citep{r29} from peer review comments. Kang et al. \citep{r30} were pioneers in introducing the first large-scale peer review dataset, PeerRead, comprising full-text papers from ACL, ICLR, and NIPS, along with the corresponding editorial decisions of acceptance or rejection. Based on this dataset, they also proposed two new NLP tasks and provided a simple baseline model. Building upon this dataset, Ghosal et al. \citep{r31} utilized sentiment analysis of review texts to recommend peer review decisions and achieved significant performance improvement over the baselines. \\
\indent As more conferences, journals, and platforms like NeurIPS\footnote{\href{https://neurips.cc/Conferences/2022/CallForPapers}{https://neurips.cc/Conferences/2022/CallForPapers}} , \textit{eLife}\footnote{\href{https://reviewer.elifesciences.org/author-guide/editorial-process}{https://reviewer.elifesciences.org/author-guide/editorial-process}} , \textit{PeerJ}\footnote{\href{https://peerj.com/benefits/review-history-and-peer-review/}{https://peerj.com/benefits/review-history-and-peer-review/}}, OpenReview\footnote{\href{https://openreview.net/}{https://openreview.net/}} , and \textit{F1000Research}\footnote{\href{https://f1000research.com/about}{https://f1000research.com/about}} provide openly accessible peer reviews, numerous datasets \citep{r32,r33,r34} have been constructed to explore various aspects of peer review. Lin et al. \citep{r32} proposed MOPRD, a multidisciplinary open peer review dataset for automatic generation of structured peer reviews, providing interdisciplinary research for peer review. Ghosal et al. \citep{r33} proposed a multi-layer dataset that can determine the paper section correspondence, paper aspect category, review functionality, and review significance at the sentence level, providing indicators for the comprehensiveness of peer review. Kumar et al. \citep{r34} introduced ContraSciView, a comprehensive review-pair contradiction dataset and a novel task of automatically identifying contradictions among reviewers on a given article. Ghosal et al. \citep{r35} introduced HedgePeer to investigate uncertainty in peer review and provided several baselines to predict hedge cues and spans, contributing convenience for uncertainty research in peer review. Yuan et al. \citep{r36} proposed ASAP-Review for generating peer reviews automatically. They provided automatic recognition of peer-reviewed sentence aspect based on this dataset and achieved excellent performance. Gao et al. \citep{r37} presented ACL-18 Numerical, focusing on predicting changes in rebuttal scores post-review. Their results indicate that the final scores of reviewers are significantly influenced by both their initial scores and the initial scores of other reviewers, emphasizing the interplay between these factors. Dycke et al. \citep{r38} compiled a multi-domain corpus, NLPEER. They established a unified data representation and enhanced previous peer-reviewed datasets. Additionally, datasets for meta-review generation \citep{r39}, discourse structure analysis of review opinions \citep{r40}, and process analysis of peer review \citep{r41} have been introduced for diverse research purposes. The aforementioned datasets have been thoroughly investigated in various aspects of peer review. Although some of them address the prediction of paper decisions, their exploration of predicting the novelty of papers and the specific contributions of peer review data to the assessment of paper novelty remains limited. Meanwhile, the aforementioned studies did not consider the crucial data of novelty scores, provided by expert reviewers at ICLR 2022.

\subsection{Methods for Evaluating Novelty of Academic Papers}
Before introducing the content of this section, it is essential to define novelty and originality conceptually. \\
\indent Novelty has been widely applied and defined from various perspectives. Boudreau et al. \citep{r63} view novelty as the ability to recombine existing knowledge components in unprecedented ways, contributing to incremental progress in a specific field. Foster et al. \citep{r64} define novelty as the difference in the number of new scientific papers compared to existing ones. Arts et al. \citep{r65} define novelty as the uniqueness of specific knowledge elements. If a scientific paper contains new knowledge elements, it indicates that the paper provides novel information. While scholars offer different interpretations of novelty in scientific papers, they all agree that novelty refers to the quality of presenting new information within these papers \citep{r66}.\\
\indent Originality has also been defined in various ways by researchers. Hou et al. \citep{r67} argue that originality arises from deviations in existing knowledge, leading to the generation of new ideas, methods, conclusions, valuable outputs, or the promotion of further innovation. Shibayama and Wang \citep{r68} define originality as a scientific discovery that provides unique knowledge for subsequent research, knowledge that previous generations could not access. Huang et al. \citep{r69} view originality as a prerequisite for creation and innovation, emphasizing the generation of new knowledge that deviates from existing research paradigms and stimulates further innovation. While originality highlights the uniqueness and creativity of a creation or research, novelty focuses on the degree of innovation within the existing field and the emergence of new knowledge or discoveries. In most cases, originality and novelty can often be used interchangeably \citep{r5,r68}.\\
\indent In conclusion, novelty and originality are closely related in meaning. Although they have distinct definitions, in certain contexts, they can be considered synonymous.\\
\indent Presently, the majority of studies predominantly assess novelty through the references or citation. Dahlin and Behrens \citep{r3} introduced a conceptualization of radical inventions and formulated a measurement methodology to assess the radicalness of inventions, applicable not only to patents but also to academic papers. Their approach involves gauging novelty through quantified citation similarity within the context of backward citations. Subsequently, Matsumoto et al. \citep{r4} expanded upon the aforementioned measurement method, extending its application to a broader-scale analysis of novelty. Their approach exclusively relies on bibliometric data to quantify the novelty of papers across diverse research fields, countries, and temporal contexts. Uzzi et al. \citep{r5} assessed 17.9 million scientific papers derived from the Web of Sciences to investigate the relationship between the reference composition of a paper and its citation count. Shibayama et al. \citep{r42} devised a method that integrates both citation data and the content of a paper. Within their framework, the determination of a paper's novelty is facilitated through the quantification of the semantic distance among references. Luo et al. \citep{r43} introduce a new approach to measuring the novelty of papers from the perspective of question-method combination, they demonstrated the effectiveness of the proposed methods by case studies and statistical analysis. Yin et al. \citep{r44} measured element novelty by designing a machine learning model to calculate the word embedding similarity between the current document and previous documents.\\
\indent In addition, Park and Simoff \citep{r45} presented a novelty identification approach grounded in a generative model. Within their methodology, the assessment of a paper's novelty is contingent upon the likelihood of it being machine-generated. This determination relies on the premise that a paper exhibiting substantial similarity to machine-generated counterparts is less likely to represent original research. Amplayo et al. \citep{r46} introduced a novelty detection model based on graphs. Their model employs authors, documents, keywords, topics, and words as feature representations to construct diverse graphs. The introduction of distinct papers to these graphs induces varied modifications, and a paper inducing substantial changes is deemed to possess greater novelty. However, the aforementioned approaches either solely rely on references or overlook the utilization of peer-review reports, which are crucial for evaluating novelty. Peer-review reports serve as significant references in assessing novelty, given the substantial expertise possessed by human expert reviewers. And the content of the paper itself is also an important part of novelty evaluation. These studies have placed more emphasis on the degree of novelty rather than differentiating between the types of novelty.\\
\indent With the impressive capabilities demonstrated by LLMs in the field of Natural Language Processing (NLP), researchers have initiated investigations into harnessing LLMs for providing peer-review feedback on academic papers. Liang et al. \citep{r20} conducted extensive empirical studies to determine the utility of LLMs in offering valuable feedback for research papers. The results indicate that LLMs and human feedback are mutually complementary. Carabantes et al. \citep{r47} explored the realm of AI-assisted peer review, finding that AI-assisted peer review is feasible to a certain extent but is subject to contextual window limitations.\\
\indent Rencently, Zhao and Zhang \citep{r72} systematically analyzes scientific novelty measurements in academic papers, comparing related concepts, categorizing existing measures by data type, examining validation methods, and highlighting future research directions. They highlight that the novelty of a scientific paper is contextual information-dependent, influenced by factors such as disciplinary background, the knowledge base used for comparison, and other factors. We argue that such contextual information is implicitly embedded in both peer review texts and the original manuscript, and that LLMs, trained on vast bodies of knowledge, can assist in providing context relevant to assessing novelty. Building upon the insights gained from these studies and survey, we embarked on an exploration of integrating human knowledge with the knowledge derived from LLMs for the assessment of novelty. Extracting content related to novelty from peer-review comments and information from the methodology sections of academic papers, we utilized these data to train deep learning models for predicting method novelty. This investigation aimed to determine the effectiveness of knowledge fusion. Additionally, we devised a dedicated Knowledge-Guided module to integrate and guide the fusion of human and LLMs knowledge.

\section{Dataset}
In this section, we will introduce how our data was collected and its sources. Subsequently, we will elaborate on the preprocessing steps applied to the acquired data for further research.

\subsection{Dataset collection}
We obtain our peer review data from the OpenReview platform\footnote{\href{https://openreview.net/about}{https://openreview.net/about}}. The International Conference on Learning Representations (ICLR) is a premier conference in the field of machine learning. We wrote a web crawler code to retrieve a total of 3376 ICLR papers, each containing peer review comments and its decision, accepted (ACC), rejected (REJ), and withdrawn (WDR).
\begin{figure*}[h]%
	\centering
	\includegraphics[width=0.7\textwidth]{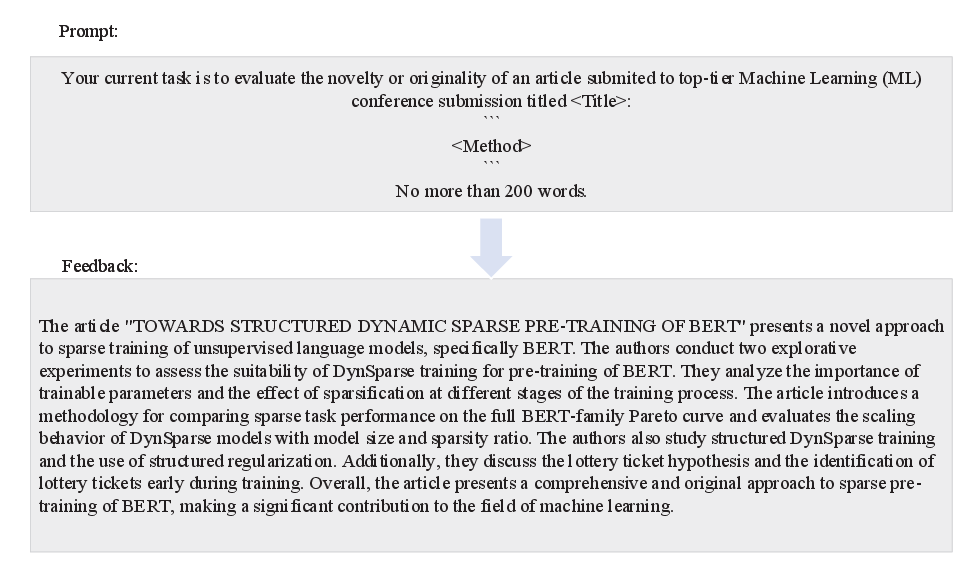}
	\vspace{-0.4cm}
	\caption{\centering{An example of our prompt and feedback.}}
	\label{fig:3}
\end{figure*}
\subsection{Dataset Pre-processing}
Firstly, we need to extract the methodology section of each paper. We utilized the GROBID (2008-2022)\footnote{\href{https://github.com/kermitt2/grobid}{https://github.com/kermitt2/grobid}} and S2ORC \citep{r48} tools to parse the PDFs into JSON format. Subsequently, we manually defined numerous rules to extract the title and the methodology sections from the parsed PDFs. For example, it is necessary to determine whether the method section appears after or before the related work, as the placement of the research method section is not fixed. To meet our research requirements, we utilized the extracted title and methodology section of the papers to construct the prompt for ChatGPT. We requested ChatGPT to provide evaluations and summaries of the novelty of the methodology section, with a constraint of staying within 200 words. The feedback for each paper was generated by ChatGPT in a single pass. An example is illustrated in Figure \ref{fig:3}. In Figure \ref{fig:3}, the “$\textless$Title$\textgreater$” in the prompt section is the paper title used to generate feedback, and “$\textless$Method$\textgreater$” represents the methodology section extracted from the paper using manually defined rules. To ensure concise feedback generation, we set a limit of no more than 200 words for the feedback, equivalent to the content of a summary. Through our manual assessment, the feedback generated by ChatGPT is relatively reliable. Currently, there is a high level of recognition in the academic community regarding its summarization capabilities. The version of ChatGPT we used is gpt-3.5-turbo. To reduce the diversity of the generated results, we set the temperature parameter to 0, while keeping all other parameters at their default settings.\\
\indent Secondly, we need to process the corresponding peer review reports. Yuan et al. \citep{r36} followed the Annual Meeting of the Association for Computational Linguistics (ACL) review guidelines and utilized human annotators to label 1000 instances for aspects such as summarization, motivation, novelty, soundness, significance, replicability, meaningful comparisons, and clarity. They trained a deep learning model for prediction and designed heuristic rules to optimize the results, concluding with a manual evaluation of the outcomes. The details of this model can be referenced in the original paper. Hence, we utilized the model to annotate peer review texts and extracted sentences marked as aspects of novelty.\\
\begin{table}[h]
	\caption{\centering{Data statistics. TNS is Technical Novelty and Significance score.}}\label{tab1}%
	\centering
	\begin{tabular}{@{}m{30mm}<{\centering} m{18mm}<{\centering}m{18mm}<{\centering}m{18mm}<{\centering}m{18mm}<{\centering} @{}}
		\toprule
		\diagbox{\textbf{TNS}}{\textbf{Decision}} & \textbf{ACC} & \textbf{REJ} & \textbf{WDR} & \textbf{Total} \\
		\midrule
		\# TNS=1    & 0 & 29 & 22 & 51\\
		\# TNS=2    & 195 & 737 & 442 & 1374 \\
		\# TNS=3    & 535 & 306 & 95  & 936 \\
		\# TNS=4    & 63 & 7 & 1  & 71 \\
		\# Papers  & 793  & 1079 & 560 & 2432\\
		\bottomrule 
	\end{tabular}
\end{table}
\indent The peer review reports for International Conference on Learning Representations (ICLR) 2022 include the review text, Correctness, Technical Novelty and Significance, Empirical Novelty and Significance, Recommendation, and Confidence. Among these, the distribution of Technical Novelty and Significance scores ranges from 1 to 4, specific descriptions presented in Appendix \ref{app2} Table \ref{tabs1}, denoted as 1: The contributions are neither significant nor novel, 2: The contributions are only marginally significant or novel, 3: The contributions are significant and somewhat new. Aspects of the contributions exist in prior work and 4: The contributions are significant, and do not exist in prior works, respectively. As we need to evaluate the novelty of the methodology section of academic papers, we retained only the scores for Technical Novelty and Significance (TNS) as the gold standard for novelty. Each paper receives evaluations from three to four reviewers, and each reviewer assigns a Technical Novelty and Significance (TNS) score. To obtain labels suitable for model training, we aggregate the scores provided by each reviewer by summing them up and then dividing by the total number of reviewers. However, discrepancies may arise when multiple reviewers have differing opinions, such as one assigning a score of 4 and another a score of 1. To address this issue, we identified the maximum and minimum scores in each review report and exclude instances where the difference exceeds 1, signifying reviewer disagreement. These cases are subsequently removed from the dataset. Ultimately, the data obtained is presented as shown in Table \ref{tab1}. From the table, it can be observed that the counts for scores 1 and 4 are relatively low. If it is predicted directly, the extremely imbalanced distribution of labels may lead to unfair prediction results. Therefore, in order to avoid this issue, we grouped scores 1 and 2 into one category, and scores 3 and 4 into another category. Based on the descriptions associated with each score, the novelty of the methods is classified into two categories: 'Low Novelty' and 'High Novelty'. Based on the descriptions of the various TNS scores the data processing method of sentiment analysis of online review \citep{r71}, we have defined 'Low Novelty' and 'High Novelty', as follow:\\
\indent \textbf{Low Novelty} refer to the methods mentioned in the paper lack innovation, are repetitive of existing work, or are merely simple extensions of existing research.\\
\indent \textbf{High Novelty} refers to the methods mentioned in the paper being innovative, unique, or involving the exploration of unknown areas.\\
\indent Following all the aforementioned preprocessing steps, the final dataset consists of 2432 instances.
\begin{figure*}[t]%
	\centering
	\includegraphics[width=1\textwidth]{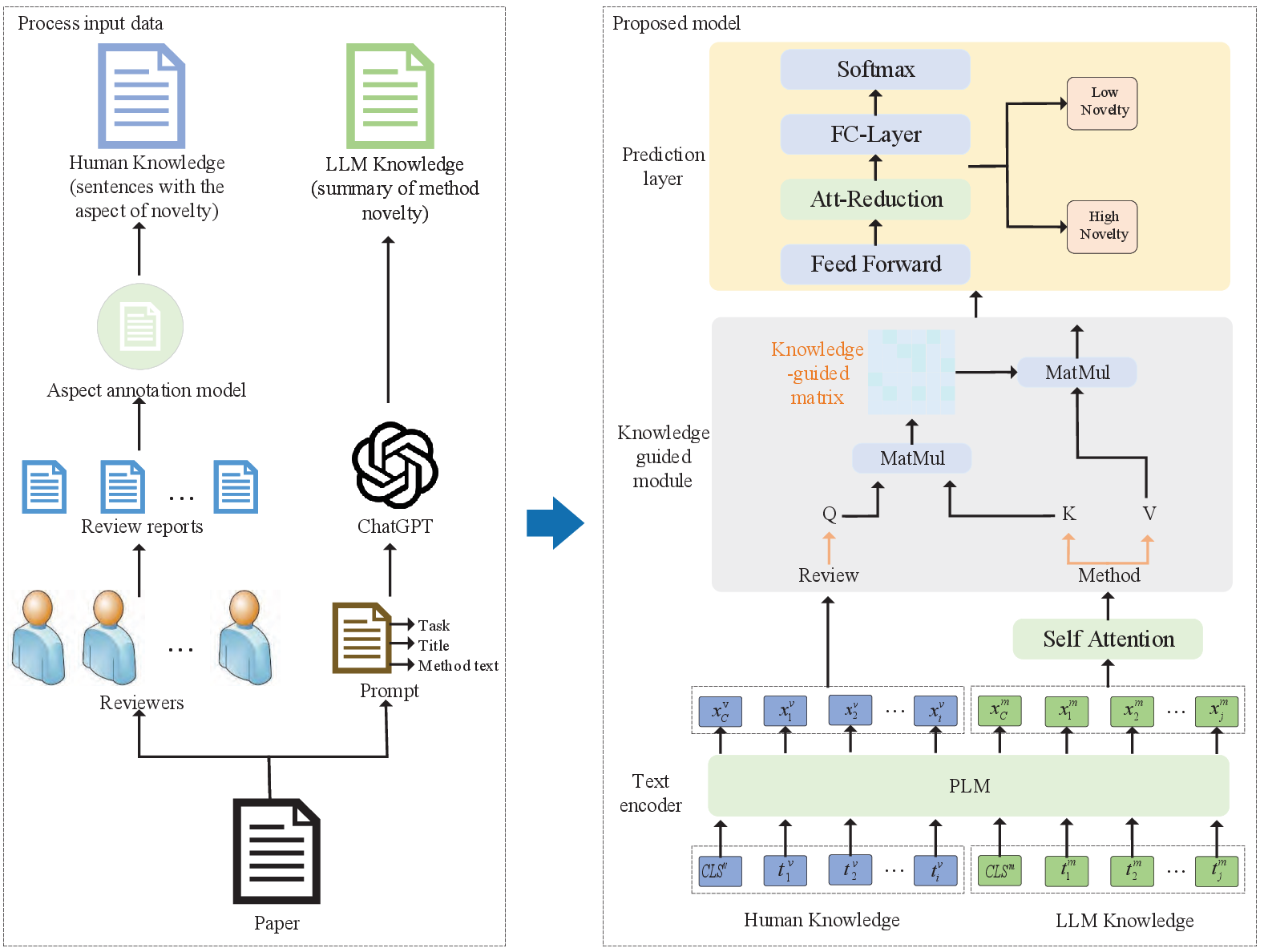}
	\vspace{-1.0cm}
	\begin{flushleft}
		{\footnotesize \textbf{Note}: Att-Reduction is Self-Attention Reduction module. PLM is pretrained language model.} 
	\end{flushleft}
	\vspace{-0.7cm}
	\caption{\centering{Overview of our study.}}
	\label{fig:4}
\end{figure*}
\section{Methodology}
In this section, we will introduce the Task definition, followed by an introduction to the Proposed Prediction model. Furthermore, as shown in Figure \ref{fig:4}, we illustrate the pre-processing process of the framework's input data on the left side and present our proposed framework (introduce in section \ref{sec4.2}), composed of a knowledge-guided fusion module, on the right side of the figure, which is applied to method novelty prediction. Then, we briefly describe how the input data is obtained. First, we need to collect all review reports (3-5 reports) for the paper whose method novelty is to be predicted. Using an aspect annotation model, we extract sentences with the aspect of novelty from all the review reports as the human knowledge input to the knowledge interaction model. Then, we replace the method text and title of the paper in the prompt shown in Figure \ref{fig:3}. This prompt is used to generate evaluation and summary of method novelty via ChatGPT as LLM knowledge.

\subsection{Task definition}
Based on the collected dataset, we defined a new task: method novelty prediction (MNP). In the context of the ICLR data we used, the novelty of the method is a key aspect that requires evaluation, which led us to define the method novelty prediction task. In addition, Leahey et al. \citep{r62} also noted in their study that one of the most common type of novelty is method novelty. Formally, given the human knowledge (sentences with the aspect of novelty) $V=(s_{1}^{v},s_{2}^{v},...,s_{i}^{v})$ consisting of $i$ sentences and LLM Knowledge (evaluation and summary of method novelty) $M=(s_{1}^{m},s_{2}^{m},...,s_{j}^{m})$ consisting of $j$ sentences. The goal of MNP is to develop a classification model $f$ that assigns predefined novelty (Low novelty and High novelty) based on the review text and feedback.

\subsection{Knowledge Interaction Model}
\label{sec4.2}
\subsubsection{Text Encoder}
First, we introduce how to obtain the embedded representation of text as input. We use the pretrained language models (PLMs, such as BERT \citep{r49}, SciBERT \citep{r50} etc.) as the text encoder to obtain the text representation. As shown in Figure \ref{fig:4}, given the sequence of review text $T^v={CLS^v,t_{1}^{v},t_{2}^{v},...,t_{i}^{v}}$ and the feedback $T^m={CLS^m,t_{1}^{m},t_{2}^{m},...,t_{j}^{m}}$ , where $i$ and $j$ is the number of text length, the output of the PLM model can be defined as:	
\begin{equation}
	X^v={x^{v}_{C},x^{v}_{1},x^{v}_{2},...,x^{v}_{i}}=PLM(T^{v};\theta^{PLM})
\end{equation}

\begin{equation}
	X^m={x^{m}_{C},x^{m}_{1},x^{m}_{2},...,x^{m}_{j}}=PLM(T^{m};\theta^{PLM})
\end{equation}
Where $x^{v}_{C}$ and $x^{m}_{C}$ is the embedding of the CLS token,  $\theta^{PLM}$ denotes the parameters of the PLM.
\subsubsection{Knowledge Guided Module}
The purpose of the knowledge guided module is to acquire knowledge from both human and artificial intelligence perspectives, and integrate their knowledge to form a fusion knowledge for prediction. The Review and the Method each contain knowledge from both human and artificial intelligence perspectives. Inspired by Wei et al. \citep{r51}, we designed a knowledge-guided module to extract and integrate this knowledge, mutually guiding them for the final predictions. First, we feed the extracted Method features $X^{m}$ into self-attention to generate method-aware feature $\alpha^{m}$. We perform linear transformations on both $X^{v}$ and $\alpha^{m}$ to obtain the query $q^{v}$ for $X^{v}$ and the key $k^{m}$ and value $v^{m}$ for $\alpha^{m}$. Then, we input $q^{v}$, $k^{m}$ and $v^{m}$ to a Sparse-Attention layer \citep{r52} to obtain knowledge-guided review sparse features $X^{vm}$:
\begin{equation}
	\alpha^{vm}=v^{m}softmax(SparseAttention(\frac{q^{v}k^{m}}{\sqrt{d^{h}}}))
\end{equation}

\begin{equation}
	X^{vm}=FC(\alpha^{vm})+X^{v}
\end{equation}
\indent Where $FC(\cdot)$ is fully connected layer. The utilization of sparse attention in this context aims to focus on method features using review characteristics, thereby obtaining features guided by the reviews. Finally, we apply a feed-forward network (FFN) for $X^{vm}$ to produce the output $S$. 
\subsubsection{Prediction Layer}
In order to conduct predictions on method novelty, it is necessary to perform dimensionality reduction on the feature $S$. Here, we design a Self-Attention Reduction module to perform dimensionality reduction on the feature $S$:
\begin{equation}
	p=\sum S*softmax(Linear(S))
\end{equation}
\indent Then we input $p$ to FC-Layer with GELU \citep{r53} activation to obtain the vector $m$. This is followed by $softmax$ activation:
\begin{equation}
	c=softmax(FC(p))
\end{equation}
\indent Where output $c$ is the final prediction. For the MNP task, given its binary nature, we employ negative sampling techniques \citep{r54} to compute the loss between predictions and true values and update the model parameters.

\section{Experiments}
In this section, we first introduce the experimental setup and baselines, then we report the experimental results, conduct analysis and the ablation study analysis. Finally, we conducted a case study to demonstrate the effectiveness of our method.
\subsection{Implementation Details}
All models are run with V100 or RTX 4090 GPU. For feature extraction, we used pre-trained language model as textual encoder. We mainly chose several pre-trained language models such as BERT \citep{r49}, RoBERTa \citep{r55}, SciBERT \citep{r50}, XLNet \citep{r56} and ALBERT \citep{r57}. The dimension of pre-trained language text embeddings is 768 by default. Number of output linear layers for NMP task is 2. We randomly select 5 negative samples for the NMP task during training in main experiments. We use Adam \citep{r58} with an initial learning rate of 1e-5 and update parameters with a batch size of 16. The trained model represents the best-performing model based on the validation set results. We split our dataset by a ratio of 8:1:1 for training, development and testing. Number of Self Attention head and Sparse Attention head is 12 and 6. The sparse dimension is set as 128. The dropout is set as 0.2. We adopt Accuracy and Weighted $F_1$ as the evaluation metrics for our dataset to evaluate the performance of the model.
\subsection{Baseline Models}
In the realm of natural language processing, several state-of-the-art language models have emerged as foundational baselines, each contributing to the advancement of various downstream tasks. We selected the following five PLMs and commonly used LLMs as baselines to fully validate the performance of our method.\\
\indent \textbf{BERT} \citep{r49} (Bidirectional Encoder Representations from Transformers) pioneered a revolutionary approach by leveraging bidirectional context information during pre-training, leading to remarkable improvements in language understanding. \\
\indent \textbf{RoBERTa} \citep{r55} (Robustly optimized BERT approach) further enhanced BERT's performance through careful modifications in training objectives and hyperparameters, yielding robust representations.\\
\indent \textbf{SciBERT} \citep{r50}, specifically tailored for scientific literature, demonstrated the importance of domain-specific pre-training.\\
\indent \textbf{XLNet} \citep{r56}, an autoregressive model, introduced permutation language modeling, allowing it to consider all possible word permutations during training.\\
\indent \textbf{ALBERT} \citep{r57} (A Lite BERT) addressed computational efficiency concerns associated with BERT by reducing its model size while preserving its performance.\\
\indent \textbf{Llama 3.1 \citep{r60}} continues the progress of the Llama series, which has established itself as a competitive open-source model in areas such as general knowledge, steerability, math, tool use, and multilingual translation, rivaling top AI models.\\
\indent \textbf{ChatGPT and GPT-4o \citep{r12}} are LLMs developed by OpenAI based on GPT technology, featuring powerful natural language processing capabilities. As a subsequent version, GPT-4o further enhances the intelligence and processing efficiency built upon the advantages of its predecessor.\\
\indent \textbf{Claude \citep{r70}} is a series of high-performance and intelligent AI models developed by Anthropic. Claude is powerful, scalable, and one of the most trusted and reliable AI systems available today.\\
\indent It is important to note that all the pre-trained models we employed are based versions. Llama 3.1 is 8B version, ChatGPT's version is gpt-3.5-turbo-0125, GPT-4o's version is gpt-4o, Claude's version is claude-3-5-sonnet-all. When calling the API, all parameters are default parameters. The experimental results of the LLMs are the averages obtained from three rounds of experiments conducted under zero-shot conditions. The prompts for LLMs are shown in Figure \ref{fig:s4}, \ref{fig:s5} and \ref{fig:s6} in the Appendix \ref{app3}.
\subsection{Results and Analysis}
\label{sec5.3}
In this section, we use the human knowledge (HK), LLM knowledge (LLMK), and method text (MT) from Figure \ref{fig:4}, either individually or in combination, as input to baseline models and compare them with our proposed knowledge interaction model. The goal is to validate the effectiveness of the interaction between HK and LLMK in the task of method novelty prediction. In Section \ref{subsec5.3.1}, we present the results where combinations of HK, LLMK, and MT are used as inputs to the baseline models and our proposed model. In Section \ref{subsec5.3.2}, we provide the results where each of these inputs is used individually as input to the baseline models.
\subsubsection{Results of Method Novelty Prediction Using Combined Knowledge Sources}
\label{subsec5.3.1}
\begin{table}[h]
	\centering
	\caption{\centering{Results of method novelty prediction with multi-information.}} 
	\label{tab2}
	\begin{tabular}{@{}m{25mm}<{\centering}m{20mm}<{\centering}m{20mm}<{\centering}m{20mm}<{\centering}m{20mm}<{\centering}@{}}
		\toprule
		\multirow{2}{*}{\textbf{Models}}&\multicolumn{2}{c}{\textbf{HK \& MT}} &\multicolumn{2}{c}{\textbf{HK \& LLMK}}\\
		&\bm{$F_1$}&\textbf{Acc}&\bm{$F_1$}&\textbf{Acc}\\
		\midrule
		
		BERT & 0.70 & 0.71 & 0.72 & 0.72 \\
		RoBERTa & 0.68 & 0.67 & 0.71 & 0.71 \\
		SciBERT & 0.70 & 0.71 & 0.73 & 0.74 \\
		XLNet & 0.62 & 0.62 & 0.69 & 0.71 \\
		ALBERT & 0.52 & 0.54 & 0.64 & 0.64 \\
		LLama 3.1 & 0.31 & 0.43 & 0.44 & 0.50 \\
		ChatGPT & 0.68 & 0.68 & 0.59 & 0.60 \\
		GPT-4o & 0.73 & 0.73 & 0.68 & 0.68 \\
		Claude & 0.68 & 0.67 & 0.69 & 0.68 \\
		Ours-BERT & - & - & 0.81 & 0.82 \\
		Ours-RoBERTa & - & - & 0.78 & 0.79 \\
		Ours-SciBERT & - & - & \textbf{0.83} & \textbf{0.84} \\
		Ours-XLNet & - & - & 0.78 & 0.79 \\
		Ours-ALBERT & - & - & 0.76 & 0.77 \\
		\bottomrule
	\end{tabular}
	\begin{tablenotes}
		\footnotesize
		\item \textbf{Note:} “-” means this method does not take this into account as an input. The results of the LLMs are the averages obtained from three rounds of testing.
	\end{tablenotes}
\end{table}
The results are illustrated in Table \ref{tab2}. In this table, "Ours-BERT" signifies that the text embedding layer in our proposed model is based on BERT, maintaining analogous configurations for other components. Among the models that use HK and Method as inputs, BERT and SciBERT have outstanding performance, while other models have poor performance, which may be related to their pre training methods and training texts. Evidently, in comparison with the HK and MT as input, under the assistance of LLMs knowledge (i.e., both peer reviews and feedback used as inputs), the deep learning model demonstrates superior performance. SciBERT achieved an accuracy of 0.74 and an F1 value of 0.73, with a better improvement compared to models with HK and MT inputs. The performance of other models has also increased to varying degrees. This indicates that LLMs can effectively summarize the essential content of the method section, and when combined with human reviewers' perspectives, enhance the overall performance of the model. We analysis the reason is the concise feedback provided by ChatGPT encompasses the core content of the method section of academic papers, facilitating a more accessible understanding for the model. As observed from the Table \ref{tab2}, SciBERT, serving as a baseline, exhibits the best performance. This can be attributed to the fact that SciBERT's training data is derived from academic texts, making it more conducive to the task of method novelty prediction. Furthermore, the results reveal that the input of both HK and LLMK enhances the performance of the five PLMs to varying extents. Notably, ALBERT exhibits the most substantial improvement, indicating that the combination of LLMs with human reviewers is beneficial across different parameter scales of models. Therefore, the synthesized data of human and LLM can lead to more effective model.\\
\indent In addition, we present the results of our proposed method (HK and LLMK as the input). As shown in Table \ref{tab2}, the model based on SciBERT performs the best, while the other models also demonstrate relative competitiveness. Compared to the baseline models, our proposed method exhibits the best performance, indicating its effectiveness in integrating human knowledge and knowledge from LLMs. It can guide the deep learning model to comprehend knowledge effectively. In the model with HK and LLMK as inputs, the model with added knowledge guided module has shown varying degrees of improvement compared to the baseline model. This indicates that our designed knowledge guided module enables the model to successfully learn and integrate knowledge from humans and LLM, and can be effectively used for final prediction. In addition, with the support of its massive knowledge, the content generated by LLM can supplement the content provided by humans, further stimulating the potential of deep learning models, enabling them to learn the most relevant knowledge for method novelty prediction.\\ 
\indent Finally, we separately input two types of text combinations into the LLMs to test the effectiveness of human fusion of knowledge from the LLMs. The results show that the smallest parameter model, LLama 3.1, performs very poorly, while other LLMs achieve results close to their true capabilities. Under zero-shot conditions, these models remain competitive with PLMs that have undergone multiple rounds of training. When using different inputs, only LLama 3.1 shows performance improvement when HK and LLMK are used as inputs, whereas other LLMs either experience a performance decline or remain unaffected. This indicates that integrating knowledge from LLMs and humans is effective for smaller-parameter models but has little impact on models with massive parameters. Furthermore, we observe that our proposed method significantly outperforms LLMs, even though the performance of these models, such as GPT-4o, is already outstanding. This underscores the effectiveness of our proposed knowledge fusion framework and highlights that the collaboration between LLMs and human knowledge can enhance the performance of smaller-parameter models.
\subsubsection{Results of Method Novelty Prediction Using Individual Knowledge Sources}
\label{subsec5.3.2}
\begin{table}[h]
	\centering
	\caption{\centering{Results of method novelty prediction with specific inputs.}} 
	\label{tab3}
	\begin{tabular}{@{}m{20mm}<{\centering}m{12mm}<{\centering}m{12mm}<{\centering}m{12mm}<{\centering}m{12mm}<{\centering}m{12mm}<{\centering}m{12mm}<{\centering}@{}}
		\toprule
		\multirow{2}{*}{\textbf{Models}}&\multicolumn{2}{c}{\textbf{HK}} &\multicolumn{2}{c}{\textbf{MT}}&\multicolumn{2}{c}{\textbf{LLMK}}\\
		&\bm{$F_1$}&\textbf{Acc}&\bm{$F_1$}&\textbf{Acc}&\bm{$F_1$}&\textbf{Acc}\\
		\midrule
		BERT&0.70&0.70&0.40&0.56&0.41&0.56\\
		RoBERTa&0.70&0.70&0.56&0.59&0.40&0.56\\
		SciBERT&\textbf{0.71}&\textbf{0.71}&0.55&0.60&0.44&0.56\\
		XLNet&0.70&0.70&0.55&0.60&0.52&0.57\\
		ALBERT&0.68&0.68&0.51&0.57&0.40&0.56\\
		LLama 3.1&0.64&0.64&0.29&0.43&-&-\\
		ChatGPT&0.67&0.67&0.30&0.44&-&-\\
		GPT-4o&0.68&0.69&0.41&0.48&-&-\\
		Claude&0.69&0.69&0.51&0.52&-&-\\
		\bottomrule
	\end{tabular}
	\begin{tablenotes}
		\footnotesize
		\item \textbf{Note:} “-” means this method does not take this into account as an input. The results of the LLMs are the averages obtained from three rounds of testing. Method is method text in Figure \ref{fig:4}.
	\end{tablenotes}
\end{table}
\noindent To further validate the effectiveness of collaboration between humans and LLMs, we conducted an additional comparative experiment. In this section, we separately used the HK, MT and LLMK from the previous section as inputs to train and prompt the PLMs and LLMs. Since the LLMK was generated by ChatGPT and is closely tied to LLMs, we did not use the LLMK as a prompt for the LLMs to generate results, as we consider this to be an inherent capability of the LLMs. The final results are shown in Table \ref{tab3}.\\
\indent From the results in the Table \ref{tab3}, it can be seen that regardless of the type of text used as input, the performance is far inferior to the collaboration between human knowledge and LLMs knowledge. Specifically, when HK is used as input, we can see that both PLMs and LLMs achieve relatively good performance. This is because HK consist of sentences related to novelty evaluation, which, as human knowledge, can guide the model to make relatively accurate judgments. However, the task in this study is to predict method novelty. Although the review texts include sentences related to novelty, they are not exclusively focused on assessing method novelty. As a result, models may develop some misunderstandings and make incorrect judgments. When LLMK is used as input, the performance of the PLMs is unsatisfactory. We attribute this to the abstract nature of the feedback generated by LLMs, which is challenging for PLMs to comprehend without the assistance of human knowledge. Consequently, the PLMs struggle to understand the specific tasks they need to accomplish. Similarly, when MT is used as input, both the PLMs and the LLMs perform poorly. We believe this is because the method section is difficult for both models to understand. While some parts of the text provide an overall description, other parts may consist of abstract formulas and complex content. This poses a significant challenge for both PLMs and LLMs.\\
\indent In conclusion, based on the results, we can conclude that the collaboration between Human knowledge and LLMs knowledge is effective, at least in the context of PLMs. This further validates the effectiveness of our proposed knowledge interaction model, demonstrating that our approach successfully captures the fused knowledge of humans and LLMs regarding method novelty and applies it to the final method novelty prediction.
\subsection{Ablation study}
\begin{table}[h]
	\centering
	\caption{\centering{Results of method novelty prediction.}} 
	\label{tab4}
	\begin{tabular}{@{}m{45mm}<{\centering}m{30mm}<{\centering}m{30mm}<{\centering}@{}}
		\toprule
		\multirow{2}{*}{\textbf{Models}} & \multicolumn{2}{c}{\textbf{HK \& LLMK}}\\
		&\bm{$F_1$}&\textbf{Accuracy}\\
		\midrule
		Ours-SciBert & 0.83 & 0.84  \\
		w/o Selfatt Reduction & 0.79 & 0.80  \\
		w/o Knowledge Guided & 0.73 & 0.74  \\
		w/o Self Attention & 0.81 & 0.82  \\
		\bottomrule
	\end{tabular}
	\begin{tablenotes}
		\footnotesize
		\item \textbf{Note:} The Selfatt Reduction is Self-Attention Reduction module.
	\end{tablenotes}
\end{table}
To better understand the effectiveness of each component of our method, we have done ablation study as shown in Table \ref{tab4}. For ease of presenting ablation results, we conducted experiments exclusively on the SciBERT-based model, given its superior performance. In addition, our method mainly targets method novelty prediction tasks, so we conducted ablation experiments on method novelty prediction. We systematically removed three components, namely, Self-Attention Reduction, Knowledge Guided, and Self Attention, to examine their individual impacts on the model. Firstly, considering the Self Attention Reduction (SAR) module, it is evident that the removal of SAR resulted in a slight impact on the model's performance. This observation suggests that the module contributes to enhancing the model's ability to capture crucial information by reducing attention weights. By focusing the model more selectively on important features and disregarding noise or irrelevant information, SAR aids in improving the model's capability to capture key information. This indicates that the SAR module helps to reduce redundant knowledge and can help the model make more accurate original predictions, enabling the model to better learn knowledge related to original predictions. We further conducted experiments by removing self-attention on ChatGPT’s feedback to assess its impact. The results indicate a decline in model performance, suggesting that self-attention enhances the model's ability to model sequential data. This indicates that self attention captures method novelty related knowledge, enabling it to provide more effective knowledge in subsequent knowledge guidance modules, effectively assisting the model in guiding and integrating knowledge, and making the knowledge learned by the model more specific. Finally, we removed the knowledge-guided module to examine its impact on model performance. Results from Table \ref{tab4} indicate a significant decrease in model performance when the knowledge-guided module is omitted. This suggests that the knowledge-guided module successfully integrates knowledge from both human and LLMs, guiding the model to better understand method novelty. The guidance and integration of knowledge is crucial for predicting method novelty, as the model integrates knowledge and advantages from human experts and LLMs, and is able to learn and transform these into the ability to predict method novelty. This indicates that our method is effective and can help the model understand knowledge from different perspectives and enhance method novelty prediction.
\subsection{Case Study}
\begin{figure*}[h]%
	\centering
	\includegraphics[width=1\textwidth]{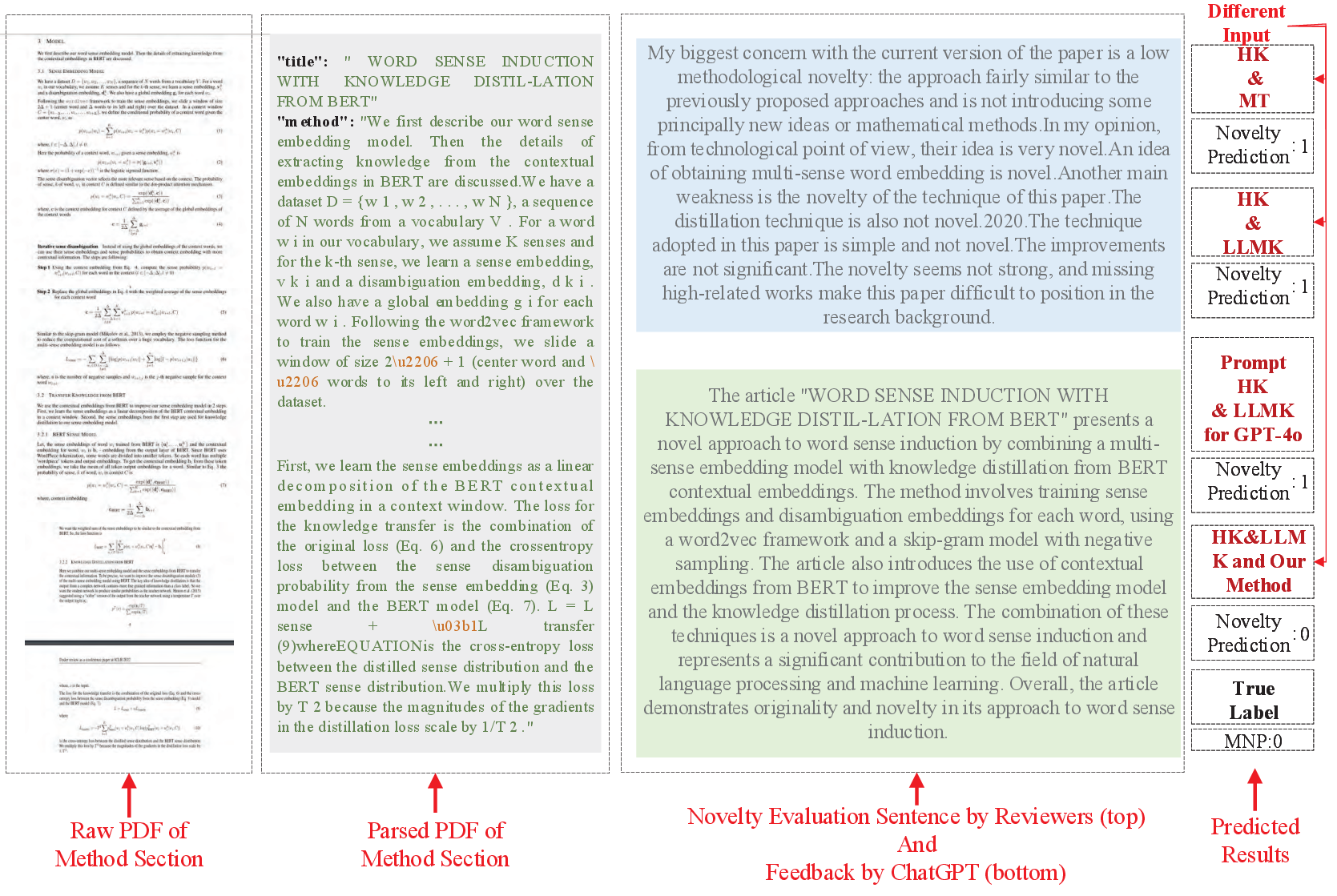}
	\vspace{-1.0cm}
	\caption{\centering{Case Study of Different Input and Our Method.}}
	\label{fig:5}
\end{figure*}
In Figure \ref{fig:5}, we conduct a case study for different input and our method. The bold text on the far right of the image means different inputs, for example, HK \& MT means taking the content with a blue background and the PDF of the directly parsed method section as inputs. Firstly, in the case of method novelty prediction, although the peer review opinions used only include evaluative sentences on novelty, there are instances where some content is perceived as original while others consider it not to be original. This poses a challenge for models relying on HK \& Method as input, resulting in errors in method novelty prediction. On the other hand, models using HK \& LLMK as input benefit from a clearer expression of the method section. However, they encounter challenges in method novelty prediction due to a lack of thorough understanding of the relationship between the review opinions and the method section, leading to incorrect predictions. In addition, we also used HK \& LLMK to prompt GPT-4o, which performed well in Section \ref{sec5.3}. However, the results show that GPT-4o failed to successfully capture the relationship between the two inputs for method novelty prediction. Finally, our proposed Knowledge Interaction Model using HK \& LLMK as input leverages attention mechanisms to assess both tasks from the perspectives of human experts and LLMs. By integrating knowledge from both viewpoints, our approach generates accurate predictions. This indicates that our approach thoroughly integrates knowledge from both human experts and LLMs, demonstrating the effectiveness of collaborative knowledge interaction between human expert knowledge and knowledge derived from LLMs.

\section{Discussion}
In this section, we will discuss the implication of our study on theoretical and practical, and limitation of our study.
\subsection{Implication}
\subsubsection{Theoretical Implication}
In this study, we collected all the paper data from ICLR 2022 along with their corresponding peer review reports. Subsequently, we extracted content related to novelty, such as review comments, and novelty scores from the peer review reports. We then utilized a PDF parsing tool to extract the method sections from the papers based on designed manual rules and evaluated the novelty of these sections using ChatGPT to obtain feedback. Building upon this foundation, we explored the integration of human and LLMs data. We conducted experiments using both the raw data and the data enhanced with feedback from ChatGPT as inputs. The results indicated that data augmented with feedback from the LLMs contributes to improved model performance. Therefore, the combination of LLMs and human-synthesized data for training deep learning models is a promising area of exploration.\\
\indent Additionally, we designed a knowledge-guided module to merge human and LLMs knowledge, guiding the deep learning model to enhance its understanding of the method novelty of papers. Extensive experiments demonstrated the effectiveness of our approach, showing its reliability. In addition, we also explored the effectiveness of the collaboration between human knowledge and LLMs knowledge. To verify this, we used human knowledge, LLMs knowledge, and the original method text as inputs separately to predict method novelty. The results indicate that relying solely on human knowledge, LLMs knowledge, or the original method text makes it challenging for the model to understand and perform the task of predicting method novelty. Moreover, the performance of these individual inputs falls significantly short compared to our designed knowledge interaction model. This further demonstrates the effectiveness of the collaboration between human and LLMs knowledge, as well as the validity of our proposed knowledge interaction design. It highlights that human-machine collaboration and the ways to optimize such collaboration could be an important direction for future research.
\subsubsection{Practical Implication}
Integrating human knowledge with the knowledge derived from LLMs proves to be effective. Recently, Microsoft introduced a novel text embedding approach \citep{r59} that leverages proprietary LLMs to generate synthetic data for various text embedding tasks across 93 languages, encompassing multiple task scenarios. This approach shares similarities with ours, as it utilizes the knowledge embedded in LLMs to serve downstream task scenarios, a perspective worth further exploration. Adopting the viewpoint of LLMs can offer more effective knowledge for current deep model training, facilitating a better understanding of tasks and their corresponding data. We believe that incorporating the knowledge from LLMs may be a worthwhile consideration for tasks involving lengthy texts.\\
\indent While the trained model can provide a preliminary assessment of the method novelty of academic papers, it is crucial to note that expert judgment remains the cornerstone for a rigorous evaluation of method novelty. Similar to our approach, machine-independent judgment of the method novelty of an academic paper has a long way to go. At present, the collaboration between humans and machines is essential, with each contributing its knowledge to mutually enhance the assessment of method novelty. \\
\indent The knowledge-guided method we propose has demonstrated promising results by integrating human and LLMs knowledge. We believe that the responsible deployment and utilization of LLMs and generative artificial intelligence may herald a transformation in research methodologies. The fusion of human knowledge with LLMs has the potential to spark significant scientific advancements, reshaping the current paradigms in various research domains and paving the way for a new landscape in scientific and technological development.
\subsection{Limitation}
Our study has several limitations that warrant emphasis. First, while our study relied on feedback from ChatGPT, more advanced models such as GPT-4 are now available and could offer enhanced insights. Additionally, our proposed model relies on feedback from a LLMs for its functioning. Furthermore, the prompts we employed in our study represent just one possible form and could benefit from further exploration to engineer more effective prompts. Although ChatGPT or GPT-4 currently have certain capabilities, they also have limitations. ChatGPT still requires certain capabilities or additional tasks, such as including current scientific databases, in order for the model to be effectively used in these experiments.\\
\indent Moreover, the knowledge-guided module we designed is an initial exploration and represents only one approach to knowledge integration; future research could investigate more sophisticated methods of knowledge fusion. Our study did not account for the impact of visual data in the methods section of academic papers, such as tables, figures, and graphs, which are essential components. Lastly, the data we used from ICLR is limited to the field of machine learning, introducing certain constraints to the generalizability of our findings.
\subsection{Practical Applicability}
Our approach leverages the summarization capabilities of LLMs and the textual evaluations of novelty found in peer reviews to automatically predict the novelty of paper method. We believe that our approach offers a degree of practicality. Although most peer review data are difficult to obtain, there are increasingly more open peer review data available, such as those from \textit{F1000Research}, \textit{eLife}, \textit{PeerJ}, \textit{Nature Communications} and \textit{PLOS ONE}, which include data from multiple fields such as biology and medicine. When applying our approach to other domains, a small amount of data can be annotated for transfer learning, knowledge distillation or model fine-tuning methods such as few-shot learning. As more peer review datasets become accessible, our method can provide a pipeline for utilizing these data. Human reviewers often provide valuable insights, but their assessments can sometimes conflict or vary in detail, leading to inconsistencies. Automated models that integrate these diverse inputs can deliver more consistent and objective evaluations of methodological novelty. Currently, most open review reports, such as those from \textit{PLOS ONE} and ICLR 2024, do not explicitly assess methodological novelty; instead, such judgments are often embedded implicitly or explicitly within the review text. In these cases, automated evaluation becomes essential to synthesize multiple reviewers' perspectives and offer readers a more objective and unified assessment. Given that readers tend to prioritize research perceived as highly novel, our approach can serve as a valuable tool for those specifically interested in identifying and exploring studies with significant methodological innovation. Moreover, our approach is scalable and can be adapted for automated evaluation of other types of novelty by replacing the method text with other relevant sections of academic papers when data related to other types of novelty becomes available. It is important to note that our aim is not to replace human judgment but to assist in secondary evaluations and provide a reference.
\section{Conclusion and Future works}
\label{con}
In this study, we propose utilizing peer reviews and the methodology section of academic papers to predict the method novelty of academic papers. To achieve this, we extract content related to novelty from peer reviews and use the methodology section of academic papers to construct prompts for input into LLMs, obtaining feedback on method novelty. Based on this data, we train a deep learning model to investigate whether the combination of human and LLMs knowledge is more advantageous for predicting method novelty. Experimental results indicate a significant improvement in the performance of deep learning models when human and LLMs knowledge are combined. Additionally, we introduce a knowledge-guided module to integrate human and LLMs knowledge and guide the deep learning model. Experimental results demonstrate the effectiveness of our approach in fusing human and LLMs knowledge, and significantly enhance the performance of deep learning models. We believe that the collaboration between humans and LLMs can bring about novel research paradigms in the scientific community.\\
\indent In the future, we plan to collect data from a broader range of disciplines and journals to assess novelty since our current dataset is limited to the field of computer science and conference papers. We plan to conduct collaborative novelty evaluations between human and LLM knowledge for other types of novelty (non-method novelty). Furthermore, we aim to explore more effective methods to integrate human and LLM knowledge, obtaining more efficient synthetic data from the collaboration between humans and LLMs. We also plan to evaluate overall novelty using combinations of different sections of the paper, such as the introduction and related work, rather than limiting the assessment to the methodology alone. Finally, utilizing retrieval-augmented techniques to further enhance the feedback mechanism of LLMs is a method we plan to explore in the future.

\section{Acknowledgment}
This work is supported by the National Natural Science Foundation of China (Grant No. 72074113) and the Graduate Research and Practice Innovation Program of Jiangsu Province (Grant No. KYCX22\_0497)
\bibliography{mybibfile}

\begin{thebibliography}{72}
\expandafter\ifx\csname natexlab\endcsname\relax\def\natexlab#1{#1}\fi
\providecommand{\url}[1]{\texttt{#1}}
\providecommand{\href}[2]{#2}
\providecommand{\path}[1]{#1}
\providecommand{\DOIprefix}{doi:}
\providecommand{\ArXivprefix}{arXiv:}
\providecommand{\URLprefix}{URL: }
\providecommand{\Pubmedprefix}{pmid:}
\providecommand{\doi}[1]{\href{http://dx.doi.org/#1}{\path{#1}}}
\providecommand{\Pubmed}[1]{\href{pmid:#1}{\path{#1}}}
\providecommand{\bibinfo}[2]{#2}
\ifx\xfnm\relax \def\xfnm[#1]{\unskip,\space#1}\fi
\bibitem[{Amplayo et~al.(2018)Amplayo, Hong \& Song}]{r46}
\bibinfo{author}{Amplayo, R.~K.}, \bibinfo{author}{Hong, S.}, \&
  \bibinfo{author}{Song, M.} (\bibinfo{year}{2018}).
\newblock \bibinfo{title}{Network-based approach to detect novelty of scholarly
  literature}.
\newblock {\it \bibinfo{journal}{Information sciences}\/},  {\it
  \bibinfo{volume}{422}\/}, \bibinfo{pages}{542--557}.
\bibitem[{Anthropic(2024)}]{r70}
\bibinfo{author}{Anthropic, A.} (\bibinfo{year}{2024}).
\newblock \bibinfo{title}{Claude 3.5 sonnet model card addendum}.
\newblock {\it \bibinfo{journal}{Claude-3.5 Model Card}\/},  {\it
  \bibinfo{volume}{3}\/}.
\bibitem[{Arts et~al.(2021)Arts, Hou \& Gomez}]{r65}
\bibinfo{author}{Arts, S.}, \bibinfo{author}{Hou, J.}, \&
  \bibinfo{author}{Gomez, J.~C.} (\bibinfo{year}{2021}).
\newblock \bibinfo{title}{Natural language processing to identify the creation
  and impact of new technologies in patent text: Code, data, and new measures}.
\newblock {\it \bibinfo{journal}{Research Policy}\/},  {\it
  \bibinfo{volume}{50}\/}, \bibinfo{pages}{104144}. \URLprefix
  \url{https://www.sciencedirect.com/science/article/pii/S0048733320302195}.
  \DOIprefix\doi{https://doi.org/10.1016/j.respol.2020.104144}.
\bibitem[{Ayers et~al.(2023)Ayers, Poliak, Dredze, Leas, Zhu, Kelley, Faix,
  Goodman, Longhurst, Hogarth et~al.}]{r13}
\bibinfo{author}{Ayers, J.~W.}, \bibinfo{author}{Poliak, A.},
  \bibinfo{author}{Dredze, M.}, \bibinfo{author}{Leas, E.~C.},
  \bibinfo{author}{Zhu, Z.}, \bibinfo{author}{Kelley, J.~B.},
  \bibinfo{author}{Faix, D.~J.}, \bibinfo{author}{Goodman, A.~M.},
  \bibinfo{author}{Longhurst, C.~A.}, \bibinfo{author}{Hogarth, M.} et~al.
  (\bibinfo{year}{2023}).
\newblock \bibinfo{title}{Comparing physician and artificial intelligence
  chatbot responses to patient questions posted to a public social media
  forum}.
\newblock {\it \bibinfo{journal}{JAMA internal medicine}\/},  {\it
  \bibinfo{volume}{183}\/}, \bibinfo{pages}{589--596}.
\bibitem[{Beltagy et~al.(2019)Beltagy, Lo \& Cohan}]{r50}
\bibinfo{author}{Beltagy, I.}, \bibinfo{author}{Lo, K.}, \&
  \bibinfo{author}{Cohan, A.} (\bibinfo{year}{2019}).
\newblock \bibinfo{title}{{S}ci{BERT}: A pretrained language model for
  scientific text}.
\newblock In \bibinfo{editor}{K.~Inui}, \bibinfo{editor}{J.~Jiang},
  \bibinfo{editor}{V.~Ng}, \& \bibinfo{editor}{X.~Wan} (Eds.), {\it
  \bibinfo{booktitle}{Proceedings of the 2019 Conference on Empirical Methods
  in Natural Language Processing and the 9th International Joint Conference on
  Natural Language Processing (EMNLP-IJCNLP)}\/} (pp.
  \bibinfo{pages}{3615--3620}).
\newblock \bibinfo{address}{Hong Kong, China}: \bibinfo{publisher}{Association
  for Computational Linguistics}.
\newblock \URLprefix \url{https://aclanthology.org/D19-1371}.
  \DOIprefix\doi{10.18653/v1/D19-1371}.
\bibitem[{Boudreau et~al.(2016)Boudreau, Guinan, Lakhani \& Riedl}]{r63}
\bibinfo{author}{Boudreau, K.~J.}, \bibinfo{author}{Guinan, E.~C.},
  \bibinfo{author}{Lakhani, K.~R.}, \& \bibinfo{author}{Riedl, C.}
  (\bibinfo{year}{2016}).
\newblock \bibinfo{title}{Looking across and looking beyond the knowledge
  frontier: Intellectual distance, novelty, and resource allocation in
  science}.
\newblock {\it \bibinfo{journal}{Manage. Sci.}\/},  {\it
  \bibinfo{volume}{62}\/}, \bibinfo{pages}{2765–2783}. \URLprefix
  \url{https://doi.org/10.1287/mnsc.2015.2285}.
  \DOIprefix\doi{10.1287/mnsc.2015.2285}.
\bibitem[{Carabantes et~al.(2023)Carabantes, Gonz{\'a}lez-Geraldo \&
  Jover}]{r47}
\bibinfo{author}{Carabantes, D.}, \bibinfo{author}{Gonz{\'a}lez-Geraldo,
  J.~L.}, \& \bibinfo{author}{Jover, G.} (\bibinfo{year}{2023}).
\newblock \bibinfo{title}{Chatgpt could be the reviewer of your next scientific
  paper. evidence on the limits of ai-assisted academic reviews}.
\newblock {\it \bibinfo{journal}{Profesional de la informaci{\'o}n/Information
  Professional}\/},  {\it \bibinfo{volume}{32}\/}.
\bibitem[{Dahlin \& Behrens(2005)}]{r3}
\bibinfo{author}{Dahlin, K.~B.}, \& \bibinfo{author}{Behrens, D.~M.}
  (\bibinfo{year}{2005}).
\newblock \bibinfo{title}{When is an invention really radical?: Defining and
  measuring technological radicalness}.
\newblock {\it \bibinfo{journal}{Research policy}\/},  {\it
  \bibinfo{volume}{34}\/}, \bibinfo{pages}{717--737}.
\bibitem[{Devlin et~al.(2019)Devlin, Chang, Lee \& Toutanova}]{r49}
\bibinfo{author}{Devlin, J.}, \bibinfo{author}{Chang, M.-W.},
  \bibinfo{author}{Lee, K.}, \& \bibinfo{author}{Toutanova, K.}
  (\bibinfo{year}{2019}).
\newblock \bibinfo{title}{{BERT}: Pre-training of deep bidirectional
  transformers for language understanding}.
\newblock In \bibinfo{editor}{J.~Burstein}, \bibinfo{editor}{C.~Doran}, \&
  \bibinfo{editor}{T.~Solorio} (Eds.), {\it \bibinfo{booktitle}{Proceedings of
  the 2019 Conference of the North {A}merican Chapter of the Association for
  Computational Linguistics: Human Language Technologies, Volume 1 (Long and
  Short Papers)}\/} (pp. \bibinfo{pages}{4171--4186}).
\newblock \bibinfo{address}{Minneapolis, Minnesota}:
  \bibinfo{publisher}{Association for Computational Linguistics}.
\newblock \URLprefix \url{https://aclanthology.org/N19-1423}.
  \DOIprefix\doi{10.18653/v1/N19-1423}.
\bibitem[{Dubey et~al.(2024)Dubey, Jauhri, Pandey, Kadian, Al-Dahle, Letman,
  Mathur, Schelten, Yang, Fan, Goyal, Hartshorn, Yang, Mitra, Sravankumar,
  Korenev, Hinsvark, Rao, Zhang, Rodriguez, Gregerson, Spataru, Roziere, Biron,
  Tang, Chern, Caucheteux, Nayak, Bi, Marra, McConnell, Keller, Touret, Wu,
  Wong, Ferrer, Nikolaidis, Allonsius, Song, Pintz, Livshits, Esiobu,
  Choudhary, Mahajan, Garcia-Olano, Perino, Hupkes, Lakomkin, AlBadawy,
  Lobanova, Dinan, Smith, Radenovic, Zhang, Synnaeve, Lee, Anderson, Nail,
  Mialon, Pang, Cucurell, Nguyen, Korevaar, Xu, Touvron, Zarov, Ibarra,
  Kloumann, Misra, Evtimov, Copet, Lee, Geffert, Vranes, Park, Mahadeokar,
  Shah, van~der Linde, Billock, Hong, Lee, Fu, Chi, Huang, Liu, Wang, Yu,
  Bitton, Spisak, Park, Rocca, Johnstun, Saxe, Jia, Alwala, Upasani, Plawiak,
  Li, Heafield et~al.}]{r60}
\bibinfo{author}{Dubey, A.}, \bibinfo{author}{Jauhri, A.},
  \bibinfo{author}{Pandey, A.}, \bibinfo{author}{Kadian, A.},
  \bibinfo{author}{Al-Dahle, A.}, \bibinfo{author}{Letman, A.},
  \bibinfo{author}{Mathur, A.}, \bibinfo{author}{Schelten, A.},
  \bibinfo{author}{Yang, A.}, \bibinfo{author}{Fan, A.},
  \bibinfo{author}{Goyal, A.}, \bibinfo{author}{Hartshorn, A.},
  \bibinfo{author}{Yang, A.}, \bibinfo{author}{Mitra, A.},
  \bibinfo{author}{Sravankumar, A.}, \bibinfo{author}{Korenev, A.},
  \bibinfo{author}{Hinsvark, A.}, \bibinfo{author}{Rao, A.},
  \bibinfo{author}{Zhang, A.}, \bibinfo{author}{Rodriguez, A.},
  \bibinfo{author}{Gregerson, A.}, \bibinfo{author}{Spataru, A.},
  \bibinfo{author}{Roziere, B.}, \bibinfo{author}{Biron, B.},
  \bibinfo{author}{Tang, B.}, \bibinfo{author}{Chern, B.},
  \bibinfo{author}{Caucheteux, C.}, \bibinfo{author}{Nayak, C.},
  \bibinfo{author}{Bi, C.}, \bibinfo{author}{Marra, C.},
  \bibinfo{author}{McConnell, C.}, \bibinfo{author}{Keller, C.},
  \bibinfo{author}{Touret, C.}, \bibinfo{author}{Wu, C.},
  \bibinfo{author}{Wong, C.}, \bibinfo{author}{Ferrer, C.~C.},
  \bibinfo{author}{Nikolaidis, C.}, \bibinfo{author}{Allonsius, D.},
  \bibinfo{author}{Song, D.}, \bibinfo{author}{Pintz, D.},
  \bibinfo{author}{Livshits, D.}, \bibinfo{author}{Esiobu, D.},
  \bibinfo{author}{Choudhary, D.}, \bibinfo{author}{Mahajan, D.},
  \bibinfo{author}{Garcia-Olano, D.}, \bibinfo{author}{Perino, D.},
  \bibinfo{author}{Hupkes, D.}, \bibinfo{author}{Lakomkin, E.},
  \bibinfo{author}{AlBadawy, E.}, \bibinfo{author}{Lobanova, E.},
  \bibinfo{author}{Dinan, E.}, \bibinfo{author}{Smith, E.~M.},
  \bibinfo{author}{Radenovic, F.}, \bibinfo{author}{Zhang, F.},
  \bibinfo{author}{Synnaeve, G.}, \bibinfo{author}{Lee, G.},
  \bibinfo{author}{Anderson, G.~L.}, \bibinfo{author}{Nail, G.},
  \bibinfo{author}{Mialon, G.}, \bibinfo{author}{Pang, G.},
  \bibinfo{author}{Cucurell, G.}, \bibinfo{author}{Nguyen, H.},
  \bibinfo{author}{Korevaar, H.}, \bibinfo{author}{Xu, H.},
  \bibinfo{author}{Touvron, H.}, \bibinfo{author}{Zarov, I.},
  \bibinfo{author}{Ibarra, I.~A.}, \bibinfo{author}{Kloumann, I.},
  \bibinfo{author}{Misra, I.}, \bibinfo{author}{Evtimov, I.},
  \bibinfo{author}{Copet, J.}, \bibinfo{author}{Lee, J.},
  \bibinfo{author}{Geffert, J.}, \bibinfo{author}{Vranes, J.},
  \bibinfo{author}{Park, J.}, \bibinfo{author}{Mahadeokar, J.},
  \bibinfo{author}{Shah, J.}, \bibinfo{author}{van~der Linde, J.},
  \bibinfo{author}{Billock, J.}, \bibinfo{author}{Hong, J.},
  \bibinfo{author}{Lee, J.}, \bibinfo{author}{Fu, J.}, \bibinfo{author}{Chi,
  J.}, \bibinfo{author}{Huang, J.}, \bibinfo{author}{Liu, J.},
  \bibinfo{author}{Wang, J.}, \bibinfo{author}{Yu, J.},
  \bibinfo{author}{Bitton, J.}, \bibinfo{author}{Spisak, J.},
  \bibinfo{author}{Park, J.}, \bibinfo{author}{Rocca, J.},
  \bibinfo{author}{Johnstun, J.}, \bibinfo{author}{Saxe, J.},
  \bibinfo{author}{Jia, J.}, \bibinfo{author}{Alwala, K.~V.},
  \bibinfo{author}{Upasani, K.}, \bibinfo{author}{Plawiak, K.},
  \bibinfo{author}{Li, K.}, \bibinfo{author}{Heafield, K.} et~al.
  (\bibinfo{year}{2024}).
\newblock \bibinfo{title}{The llama 3 herd of models}.
\newblock \URLprefix \url{https://arxiv.org/abs/2407.21783}.
  \href{http://arxiv.org/abs/2407.21783}{\tt arXiv:2407.21783}.
\bibitem[{Dycke et~al.(2023)Dycke, Kuznetsov \& Gurevych}]{r38}
\bibinfo{author}{Dycke, N.}, \bibinfo{author}{Kuznetsov, I.}, \&
  \bibinfo{author}{Gurevych, I.} (\bibinfo{year}{2023}).
\newblock \bibinfo{title}{{NLP}eer: A unified resource for the computational
  study of peer review}.
\newblock In \bibinfo{editor}{A.~Rogers}, \bibinfo{editor}{J.~Boyd-Graber}, \&
  \bibinfo{editor}{N.~Okazaki} (Eds.), {\it \bibinfo{booktitle}{Proceedings of
  the 61st Annual Meeting of the Association for Computational Linguistics
  (Volume 1: Long Papers)}\/} (pp. \bibinfo{pages}{5049--5073}).
\newblock \bibinfo{address}{Toronto, Canada}: \bibinfo{publisher}{Association
  for Computational Linguistics}.
\newblock \URLprefix \url{https://aclanthology.org/2023.acl-long.277}.
  \DOIprefix\doi{10.18653/v1/2023.acl-long.277}.
\bibitem[{Foster et~al.(2021)Foster, Shi \& Evans}]{r64}
\bibinfo{author}{Foster, J.~G.}, \bibinfo{author}{Shi, F.}, \&
  \bibinfo{author}{Evans, J.} (\bibinfo{year}{2021}).
\newblock {\it \bibinfo{title}{{Surprise! Measuring Novelty as Expectation
  Violation}}\/}.
\newblock \bibinfo{type}{SocArXiv} \bibinfo{number}{2t46f} Center for Open
  Science.
\newblock \URLprefix \url{https://ideas.repec.org/p/osf/socarx/2t46f.html}.
  \DOIprefix\doi{10.31219/osf.io/2t46f}.
\bibitem[{Gao et~al.(2019)Gao, Eger, Kuznetsov, Gurevych \& Miyao}]{r37}
\bibinfo{author}{Gao, Y.}, \bibinfo{author}{Eger, S.},
  \bibinfo{author}{Kuznetsov, I.}, \bibinfo{author}{Gurevych, I.}, \&
  \bibinfo{author}{Miyao, Y.} (\bibinfo{year}{2019}).
\newblock \bibinfo{title}{Does my rebuttal matter? insights from a major {NLP}
  conference}.
\newblock In \bibinfo{editor}{J.~Burstein}, \bibinfo{editor}{C.~Doran}, \&
  \bibinfo{editor}{T.~Solorio} (Eds.), {\it \bibinfo{booktitle}{Proceedings of
  the 2019 Conference of the North {A}merican Chapter of the Association for
  Computational Linguistics: Human Language Technologies, Volume 1 (Long and
  Short Papers)}\/} (pp. \bibinfo{pages}{1274--1290}).
\newblock \bibinfo{address}{Minneapolis, Minnesota}:
  \bibinfo{publisher}{Association for Computational Linguistics}.
\newblock \URLprefix \url{https://aclanthology.org/N19-1129}.
  \DOIprefix\doi{10.18653/v1/N19-1129}.
\bibitem[{Ghosal et~al.(2022{\natexlab{a}})Ghosal, Kumar, Bharti \&
  Ekbal}]{r33}
\bibinfo{author}{Ghosal, T.}, \bibinfo{author}{Kumar, S.},
  \bibinfo{author}{Bharti, P.~K.}, \& \bibinfo{author}{Ekbal, A.}
  (\bibinfo{year}{2022}{\natexlab{a}}).
\newblock \bibinfo{title}{Peer review analyze: A novel benchmark resource for
  computational analysis of peer reviews}.
\newblock {\it \bibinfo{journal}{Plos one}\/},  {\it \bibinfo{volume}{17}\/},
  \bibinfo{pages}{e0259238}.
\bibitem[{Ghosal et~al.(2022{\natexlab{b}})Ghosal, Varanasi \& Kordoni}]{r35}
\bibinfo{author}{Ghosal, T.}, \bibinfo{author}{Varanasi, K.~K.}, \&
  \bibinfo{author}{Kordoni, V.} (\bibinfo{year}{2022}{\natexlab{b}}).
\newblock \bibinfo{title}{Hedgepeer: A dataset for uncertainty detection in
  peer reviews}.
\newblock In {\it \bibinfo{booktitle}{Proceedings of the 22nd ACM/IEEE Joint
  Conference on Digital Libraries}\/} (pp. \bibinfo{pages}{1--5}).
\bibitem[{Ghosal et~al.(2019)Ghosal, Verma, Ekbal \& Bhattacharyya}]{r31}
\bibinfo{author}{Ghosal, T.}, \bibinfo{author}{Verma, R.},
  \bibinfo{author}{Ekbal, A.}, \& \bibinfo{author}{Bhattacharyya, P.}
  (\bibinfo{year}{2019}).
\newblock \bibinfo{title}{{D}eep{S}enti{P}eer: Harnessing sentiment in review
  texts to recommend peer review decisions}.
\newblock In \bibinfo{editor}{A.~Korhonen}, \bibinfo{editor}{D.~Traum}, \&
  \bibinfo{editor}{L.~M{\`a}rquez} (Eds.), {\it \bibinfo{booktitle}{Proceedings
  of the 57th Annual Meeting of the Association for Computational
  Linguistics}\/} (pp. \bibinfo{pages}{1120--1130}).
\newblock \bibinfo{address}{Florence, Italy}: \bibinfo{publisher}{Association
  for Computational Linguistics}.
\newblock \URLprefix \url{https://aclanthology.org/P19-1106}.
  \DOIprefix\doi{10.18653/v1/P19-1106}.
\bibitem[{Heck et~al.(2023)Heck, Lubis, Ruppik, Vukovic, Feng, Geishauser, Lin,
  van Niekerk \& Gasic}]{r10}
\bibinfo{author}{Heck, M.}, \bibinfo{author}{Lubis, N.},
  \bibinfo{author}{Ruppik, B.}, \bibinfo{author}{Vukovic, R.},
  \bibinfo{author}{Feng, S.}, \bibinfo{author}{Geishauser, C.},
  \bibinfo{author}{Lin, H.-c.}, \bibinfo{author}{van Niekerk, C.}, \&
  \bibinfo{author}{Gasic, M.} (\bibinfo{year}{2023}).
\newblock \bibinfo{title}{{C}hat{GPT} for zero-shot dialogue state tracking: A
  solution or an opportunity?}
\newblock In \bibinfo{editor}{A.~Rogers}, \bibinfo{editor}{J.~Boyd-Graber}, \&
  \bibinfo{editor}{N.~Okazaki} (Eds.), {\it \bibinfo{booktitle}{Proceedings of
  the 61st Annual Meeting of the Association for Computational Linguistics
  (Volume 2: Short Papers)}\/} (pp. \bibinfo{pages}{936--950}).
\newblock \bibinfo{address}{Toronto, Canada}: \bibinfo{publisher}{Association
  for Computational Linguistics}.
\newblock \URLprefix \url{https://aclanthology.org/2023.acl-short.81}.
  \DOIprefix\doi{10.18653/v1/2023.acl-short.81}.
\bibitem[{Hendrycks \& Gimpel(2016)}]{r53}
\bibinfo{author}{Hendrycks, D.}, \& \bibinfo{author}{Gimpel, K.}
  (\bibinfo{year}{2016}).
\newblock \bibinfo{title}{Gaussian error linear units (gelus)}.
\newblock {\it \bibinfo{journal}{arXiv preprint arXiv:1606.08415}\/}, .
\bibitem[{Hou et~al.(2022)Hou, Wang \& Li}]{r67}
\bibinfo{author}{Hou, J.}, \bibinfo{author}{Wang, D.}, \& \bibinfo{author}{Li,
  J.} (\bibinfo{year}{2022}).
\newblock \bibinfo{title}{A new method for measuring the originality of
  academic articles based on knowledge units in semantic networks}.
\newblock {\it \bibinfo{journal}{Journal of Informetrics}\/},  {\it
  \bibinfo{volume}{16}\/}, \bibinfo{pages}{101306}. \URLprefix
  \url{https://www.sciencedirect.com/science/article/pii/S175115772200058X}.
  \DOIprefix\doi{https://doi.org/10.1016/j.joi.2022.101306}.
\bibitem[{Hua et~al.(2019)Hua, Nikolov, Badugu \& Wang}]{r29}
\bibinfo{author}{Hua, X.}, \bibinfo{author}{Nikolov, M.},
  \bibinfo{author}{Badugu, N.}, \& \bibinfo{author}{Wang, L.}
  (\bibinfo{year}{2019}).
\newblock \bibinfo{title}{Argument mining for understanding peer reviews}.
\newblock In \bibinfo{editor}{J.~Burstein}, \bibinfo{editor}{C.~Doran}, \&
  \bibinfo{editor}{T.~Solorio} (Eds.), {\it \bibinfo{booktitle}{Proceedings of
  the 2019 Conference of the North {A}merican Chapter of the Association for
  Computational Linguistics: Human Language Technologies, Volume 1 (Long and
  Short Papers)}\/} (pp. \bibinfo{pages}{2131--2137}).
\newblock \bibinfo{address}{Minneapolis, Minnesota}:
  \bibinfo{publisher}{Association for Computational Linguistics}.
\newblock \URLprefix \url{https://aclanthology.org/N19-1219}.
  \DOIprefix\doi{10.18653/v1/N19-1219}.
\bibitem[{Huang et~al.(2025)Huang, Huang, Liu, Luo \& Lu}]{r69}
\bibinfo{author}{Huang, S.}, \bibinfo{author}{Huang, Y.}, \bibinfo{author}{Liu,
  Y.}, \bibinfo{author}{Luo, Z.}, \& \bibinfo{author}{Lu, W.}
  (\bibinfo{year}{2025}).
\newblock \bibinfo{title}{Are large language models qualified reviewers in
  originality evaluation?}
\newblock {\it \bibinfo{journal}{Information Processing \& Management}\/},
  {\it \bibinfo{volume}{62}\/}, \bibinfo{pages}{103973}. \URLprefix
  \url{https://www.sciencedirect.com/science/article/pii/S0306457324003327}.
  \DOIprefix\doi{https://doi.org/10.1016/j.ipm.2024.103973}.
\bibitem[{Hug(2022)}]{r26}
\bibinfo{author}{Hug, S.~E.} (\bibinfo{year}{2022}).
\newblock \bibinfo{title}{Towards theorizing peer review}.
\newblock {\it \bibinfo{journal}{Quantitative Science Studies}\/},  {\it
  \bibinfo{volume}{3}\/}, \bibinfo{pages}{815--831}.
\bibitem[{Jarrahi et~al.(2023)Jarrahi, Lutz, Boyd, Oesterlund \& Willis}]{r8}
\bibinfo{author}{Jarrahi, M.~H.}, \bibinfo{author}{Lutz, C.},
  \bibinfo{author}{Boyd, K.}, \bibinfo{author}{Oesterlund, C.}, \&
  \bibinfo{author}{Willis, M.} (\bibinfo{year}{2023}).
\newblock \bibinfo{title}{Artificial intelligence in the work context}.
\bibitem[{Kang et~al.(2018)Kang, Ammar, Dalvi, van Zuylen, Kohlmeier, Hovy \&
  Schwartz}]{r30}
\bibinfo{author}{Kang, D.}, \bibinfo{author}{Ammar, W.},
  \bibinfo{author}{Dalvi, B.}, \bibinfo{author}{van Zuylen, M.},
  \bibinfo{author}{Kohlmeier, S.}, \bibinfo{author}{Hovy, E.}, \&
  \bibinfo{author}{Schwartz, R.} (\bibinfo{year}{2018}).
\newblock \bibinfo{title}{A dataset of peer reviews ({P}eer{R}ead): Collection,
  insights and {NLP} applications}.
\newblock In \bibinfo{editor}{M.~Walker}, \bibinfo{editor}{H.~Ji}, \&
  \bibinfo{editor}{A.~Stent} (Eds.), {\it \bibinfo{booktitle}{Proceedings of
  the 2018 Conference of the North {A}merican Chapter of the Association for
  Computational Linguistics: Human Language Technologies, Volume 1 (Long
  Papers)}\/} (pp. \bibinfo{pages}{1647--1661}).
\newblock \bibinfo{address}{New Orleans, Louisiana}:
  \bibinfo{publisher}{Association for Computational Linguistics}.
\newblock \URLprefix \url{https://aclanthology.org/N18-1149}.
  \DOIprefix\doi{10.18653/v1/N18-1149}.
\bibitem[{Kennard et~al.(2022)Kennard, O{'}Gorman, Das, Sharma, Bagchi,
  Clinton, Yelugam, Zamani \& McCallum}]{r40}
\bibinfo{author}{Kennard, N.~N.}, \bibinfo{author}{O{'}Gorman, T.},
  \bibinfo{author}{Das, R.}, \bibinfo{author}{Sharma, A.},
  \bibinfo{author}{Bagchi, C.}, \bibinfo{author}{Clinton, M.},
  \bibinfo{author}{Yelugam, P.~K.}, \bibinfo{author}{Zamani, H.}, \&
  \bibinfo{author}{McCallum, A.} (\bibinfo{year}{2022}).
\newblock \bibinfo{title}{{DISAPERE}: A dataset for discourse structure in peer
  review discussions}.
\newblock In \bibinfo{editor}{M.~Carpuat}, \bibinfo{editor}{M.-C. de~Marneffe},
  \& \bibinfo{editor}{I.~V. Meza~Ruiz} (Eds.), {\it
  \bibinfo{booktitle}{Proceedings of the 2022 Conference of the North American
  Chapter of the Association for Computational Linguistics: Human Language
  Technologies}\/} (pp. \bibinfo{pages}{1234--1249}).
\newblock \bibinfo{address}{Seattle, United States}:
  \bibinfo{publisher}{Association for Computational Linguistics}.
\newblock \URLprefix \url{https://aclanthology.org/2022.naacl-main.89}.
  \DOIprefix\doi{10.18653/v1/2022.naacl-main.89}.
\bibitem[{Kingma \& Ba(2014)}]{r58}
\bibinfo{author}{Kingma, D.~P.}, \& \bibinfo{author}{Ba, J.}
  (\bibinfo{year}{2014}).
\newblock \bibinfo{title}{Adam: A method for stochastic optimization}.
\newblock {\it \bibinfo{journal}{arXiv preprint arXiv:1412.6980}\/}, .
\bibitem[{Kumar et~al.(2023)Kumar, Ghosal \& Ekbal}]{r34}
\bibinfo{author}{Kumar, S.}, \bibinfo{author}{Ghosal, T.}, \&
  \bibinfo{author}{Ekbal, A.} (\bibinfo{year}{2023}).
\newblock \bibinfo{title}{When reviewers lock horns: Finding disagreements in
  scientific peer reviews}.
\newblock In \bibinfo{editor}{H.~Bouamor}, \bibinfo{editor}{J.~Pino}, \&
  \bibinfo{editor}{K.~Bali} (Eds.), {\it \bibinfo{booktitle}{Proceedings of the
  2023 Conference on Empirical Methods in Natural Language Processing}\/} (pp.
  \bibinfo{pages}{16693--16704}).
\newblock \bibinfo{address}{Singapore}: \bibinfo{publisher}{Association for
  Computational Linguistics}.
\newblock \URLprefix \url{https://aclanthology.org/2023.emnlp-main.1038}.
  \DOIprefix\doi{10.18653/v1/2023.emnlp-main.1038}.
\bibitem[{Kung et~al.(2023)Kung, Cheatham, Medenilla, Sillos, De~Leon,
  Elepa{\~n}o, Madriaga, Aggabao, Diaz-Candido, Maningo et~al.}]{r15}
\bibinfo{author}{Kung, T.~H.}, \bibinfo{author}{Cheatham, M.},
  \bibinfo{author}{Medenilla, A.}, \bibinfo{author}{Sillos, C.},
  \bibinfo{author}{De~Leon, L.}, \bibinfo{author}{Elepa{\~n}o, C.},
  \bibinfo{author}{Madriaga, M.}, \bibinfo{author}{Aggabao, R.},
  \bibinfo{author}{Diaz-Candido, G.}, \bibinfo{author}{Maningo, J.} et~al.
  (\bibinfo{year}{2023}).
\newblock \bibinfo{title}{Performance of chatgpt on usmle: potential for
  ai-assisted medical education using large language models}.
\newblock {\it \bibinfo{journal}{PLoS digital health}\/},  {\it
  \bibinfo{volume}{2}\/}, \bibinfo{pages}{e0000198}.
\bibitem[{Kusumoto et~al.(2023)Kusumoto, Bittl, Creager, Dauerman, Lala,
  McDermott, Turco, Taqueti, Fuster \& of~the Scientific
  Publications~Committee}]{r22}
\bibinfo{author}{Kusumoto, F.~M.}, \bibinfo{author}{Bittl, J.~A.},
  \bibinfo{author}{Creager, M.~A.}, \bibinfo{author}{Dauerman, H.~L.},
  \bibinfo{author}{Lala, A.}, \bibinfo{author}{McDermott, M.~M.},
  \bibinfo{author}{Turco, J.~V.}, \bibinfo{author}{Taqueti, V.~R.},
  \bibinfo{author}{Fuster, V.}, \& \bibinfo{author}{of~the Scientific
  Publications~Committee, P. R. T.~F.} (\bibinfo{year}{2023}).
\newblock \bibinfo{title}{Challenges and controversies in peer review: Jacc
  review topic of the week}.
\newblock {\it \bibinfo{journal}{Journal of the American College of
  Cardiology}\/},  {\it \bibinfo{volume}{82}\/}, \bibinfo{pages}{2054--2062}.
\bibitem[{Lan et~al.(2019)Lan, Chen, Goodman, Gimpel, Sharma \& Soricut}]{r57}
\bibinfo{author}{Lan, Z.}, \bibinfo{author}{Chen, M.},
  \bibinfo{author}{Goodman, S.}, \bibinfo{author}{Gimpel, K.},
  \bibinfo{author}{Sharma, P.}, \& \bibinfo{author}{Soricut, R.}
  (\bibinfo{year}{2019}).
\newblock \bibinfo{title}{Albert: A lite bert for self-supervised learning of
  language representations}.
\newblock {\it \bibinfo{journal}{arXiv preprint arXiv:1909.11942}\/}, .
\bibitem[{Leahey et~al.(2023)Leahey, Lee \& Funk}]{r62}
\bibinfo{author}{Leahey, E.}, \bibinfo{author}{Lee, J.}, \&
  \bibinfo{author}{Funk, R.~J.} (\bibinfo{year}{2023}).
\newblock \bibinfo{title}{What types of novelty are most disruptive?}
\newblock {\it \bibinfo{journal}{American Sociological Review}\/},  {\it
  \bibinfo{volume}{88}\/}, \bibinfo{pages}{562--597}.
  \DOIprefix\doi{10.1177/00031224231168074}.
\bibitem[{Lee et~al.(2022)Lee, Srivastava, Hardy, Thickstun, Durmus, Paranjape,
  Gerard-Ursin, Li, Ladhak, Rong et~al.}]{r14}
\bibinfo{author}{Lee, M.}, \bibinfo{author}{Srivastava, M.},
  \bibinfo{author}{Hardy, A.}, \bibinfo{author}{Thickstun, J.},
  \bibinfo{author}{Durmus, E.}, \bibinfo{author}{Paranjape, A.},
  \bibinfo{author}{Gerard-Ursin, I.}, \bibinfo{author}{Li, X.~L.},
  \bibinfo{author}{Ladhak, F.}, \bibinfo{author}{Rong, F.} et~al.
  (\bibinfo{year}{2022}).
\newblock \bibinfo{title}{Evaluating human-language model interaction}.
\newblock {\it \bibinfo{journal}{arXiv preprint arXiv:2212.09746}\/}, .
\bibitem[{Liang et~al.(2023)Liang, Zhang, Cao, Wang, Ding, Yang, Vodrahalli,
  He, Smith, Yin et~al.}]{r20}
\bibinfo{author}{Liang, W.}, \bibinfo{author}{Zhang, Y.}, \bibinfo{author}{Cao,
  H.}, \bibinfo{author}{Wang, B.}, \bibinfo{author}{Ding, D.},
  \bibinfo{author}{Yang, X.}, \bibinfo{author}{Vodrahalli, K.},
  \bibinfo{author}{He, S.}, \bibinfo{author}{Smith, D.}, \bibinfo{author}{Yin,
  Y.} et~al. (\bibinfo{year}{2023}).
\newblock \bibinfo{title}{Can large language models provide useful feedback on
  research papers? a large-scale empirical analysis}.
\newblock {\it \bibinfo{journal}{arXiv preprint arXiv:2310.01783}\/}, .
\bibitem[{Lin et~al.(2023{\natexlab{a}})Lin, Song, Zhou, Chen \& Shi}]{r27}
\bibinfo{author}{Lin, J.}, \bibinfo{author}{Song, J.}, \bibinfo{author}{Zhou,
  Z.}, \bibinfo{author}{Chen, Y.}, \& \bibinfo{author}{Shi, X.}
  (\bibinfo{year}{2023}{\natexlab{a}}).
\newblock \bibinfo{title}{Automated scholarly paper review: Concepts,
  technologies, and challenges}.
\newblock {\it \bibinfo{journal}{Information fusion}\/},  {\it
  \bibinfo{volume}{98}\/}, \bibinfo{pages}{101830}.
\bibitem[{Lin et~al.(2023{\natexlab{b}})Lin, Song, Zhou, Chen \& Shi}]{r32}
\bibinfo{author}{Lin, J.}, \bibinfo{author}{Song, J.}, \bibinfo{author}{Zhou,
  Z.}, \bibinfo{author}{Chen, Y.}, \& \bibinfo{author}{Shi, X.}
  (\bibinfo{year}{2023}{\natexlab{b}}).
\newblock \bibinfo{title}{Moprd: A multidisciplinary open peer review dataset}.
\newblock {\it \bibinfo{journal}{Neural Computing and Applications}\/},  {\it
  \bibinfo{volume}{35}\/}, \bibinfo{pages}{24191--24206}.
\bibitem[{Liu \& Shah(2023)}]{r18}
\bibinfo{author}{Liu, R.}, \& \bibinfo{author}{Shah, N.~B.}
  (\bibinfo{year}{2023}).
\newblock \bibinfo{title}{Reviewergpt? an exploratory study on using large
  language models for paper reviewing}.
\newblock {\it \bibinfo{journal}{arXiv preprint arXiv:2306.00622}\/}, .
\bibitem[{Liu et~al.(2019)Liu, Ott, Goyal, Du, Joshi, Chen, Levy, Lewis,
  Zettlemoyer \& Stoyanov}]{r55}
\bibinfo{author}{Liu, Y.}, \bibinfo{author}{Ott, M.}, \bibinfo{author}{Goyal,
  N.}, \bibinfo{author}{Du, J.}, \bibinfo{author}{Joshi, M.},
  \bibinfo{author}{Chen, D.}, \bibinfo{author}{Levy, O.},
  \bibinfo{author}{Lewis, M.}, \bibinfo{author}{Zettlemoyer, L.}, \&
  \bibinfo{author}{Stoyanov, V.} (\bibinfo{year}{2019}).
\newblock \bibinfo{title}{Roberta: A robustly optimized bert pretraining
  approach}.
\newblock {\it \bibinfo{journal}{arXiv preprint arXiv:1907.11692}\/}, .
\bibitem[{Lo et~al.(2020)Lo, Wang, Neumann, Kinney \& Weld}]{r48}
\bibinfo{author}{Lo, K.}, \bibinfo{author}{Wang, L.~L.},
  \bibinfo{author}{Neumann, M.}, \bibinfo{author}{Kinney, R.}, \&
  \bibinfo{author}{Weld, D.} (\bibinfo{year}{2020}).
\newblock \bibinfo{title}{{S}2{ORC}: The semantic scholar open research
  corpus}.
\newblock In \bibinfo{editor}{D.~Jurafsky}, \bibinfo{editor}{J.~Chai},
  \bibinfo{editor}{N.~Schluter}, \& \bibinfo{editor}{J.~Tetreault} (Eds.), {\it
  \bibinfo{booktitle}{Proceedings of the 58th Annual Meeting of the Association
  for Computational Linguistics}\/} (pp. \bibinfo{pages}{4969--4983}).
\newblock \bibinfo{address}{Online}: \bibinfo{publisher}{Association for
  Computational Linguistics}.
\newblock \URLprefix \url{https://aclanthology.org/2020.acl-main.447}.
  \DOIprefix\doi{10.18653/v1/2020.acl-main.447}.
\bibitem[{Luo et~al.(2022)Luo, Lu, He \& Wang}]{r43}
\bibinfo{author}{Luo, Z.}, \bibinfo{author}{Lu, W.}, \bibinfo{author}{He, J.},
  \& \bibinfo{author}{Wang, Y.} (\bibinfo{year}{2022}).
\newblock \bibinfo{title}{Combination of research questions and methods: A new
  measurement of scientific novelty}.
\newblock {\it \bibinfo{journal}{Journal of Informetrics}\/},  {\it
  \bibinfo{volume}{16}\/}, \bibinfo{pages}{101282}.
\bibitem[{Martins \& Astudillo(2016)}]{r52}
\bibinfo{author}{Martins, A.}, \& \bibinfo{author}{Astudillo, R.}
  (\bibinfo{year}{2016}).
\newblock \bibinfo{title}{From softmax to sparsemax: A sparse model of
  attention and multi-label classification}.
\newblock In {\it \bibinfo{booktitle}{International conference on machine
  learning}\/} (pp. \bibinfo{pages}{1614--1623}).
\newblock \bibinfo{organization}{PMLR}.
\bibitem[{Matsui et~al.(2021)Matsui, Chen, Wang \& Ferrara}]{r41}
\bibinfo{author}{Matsui, A.}, \bibinfo{author}{Chen, E.},
  \bibinfo{author}{Wang, Y.}, \& \bibinfo{author}{Ferrara, E.}
  (\bibinfo{year}{2021}).
\newblock \bibinfo{title}{The impact of peer review on the contribution
  potential of scientific papers}.
\newblock {\it \bibinfo{journal}{PeerJ}\/},  {\it \bibinfo{volume}{9}\/},
  \bibinfo{pages}{e11999}.
\bibitem[{Matsumoto et~al.(2021)Matsumoto, Shibayama, Kang \& Igami}]{r4}
\bibinfo{author}{Matsumoto, K.}, \bibinfo{author}{Shibayama, S.},
  \bibinfo{author}{Kang, B.}, \& \bibinfo{author}{Igami, M.}
  (\bibinfo{year}{2021}).
\newblock \bibinfo{title}{Introducing a novelty indicator for scientific
  research: validating the knowledge-based combinatorial approach}.
\newblock {\it \bibinfo{journal}{Scientometrics}\/},  {\it
  \bibinfo{volume}{126}\/}, \bibinfo{pages}{6891--6915}.
\bibitem[{Mikolov et~al.(2013)Mikolov, Sutskever, Chen, Corrado \& Dean}]{r54}
\bibinfo{author}{Mikolov, T.}, \bibinfo{author}{Sutskever, I.},
  \bibinfo{author}{Chen, K.}, \bibinfo{author}{Corrado, G.~S.}, \&
  \bibinfo{author}{Dean, J.} (\bibinfo{year}{2013}).
\newblock \bibinfo{title}{Distributed representations of words and phrases and
  their compositionality}.
\newblock {\it \bibinfo{journal}{Advances in neural information processing
  systems}\/},  {\it \bibinfo{volume}{26}\/}.
\bibitem[{Mulligan et~al.(2013)Mulligan, Hall \& Raphael}]{r21}
\bibinfo{author}{Mulligan, A.}, \bibinfo{author}{Hall, L.}, \&
  \bibinfo{author}{Raphael, E.} (\bibinfo{year}{2013}).
\newblock \bibinfo{title}{Peer review in a changing world: An international
  study measuring the attitudes of researchers}.
\newblock {\it \bibinfo{journal}{Journal of the American Society for
  Information Science and Technology}\/},  {\it \bibinfo{volume}{64}\/},
  \bibinfo{pages}{132--161}.
\bibitem[{Nelson \& Winter(1982)}]{r2}
\bibinfo{author}{Nelson, R.}, \& \bibinfo{author}{Winter, S.}
  (\bibinfo{year}{1982}).
\newblock \bibinfo{title}{An evolutionary theory of economic change harvard
  university press}.
\newblock {\it \bibinfo{journal}{Cambridge, MA and London}\/}, .
\bibitem[{OpenAI(2024)}]{r12}
\bibinfo{author}{OpenAI} (\bibinfo{year}{2024}).
\newblock \bibinfo{title}{Gpt-4 technical report}, .
\newblock \URLprefix \url{https://arxiv.org/abs/2303.08774}.
  \href{http://arxiv.org/abs/2303.08774}{\tt arXiv:2303.08774}.
\bibitem[{Ouyang et~al.(2022)Ouyang, Wu, Jiang, Almeida, Wainwright, Mishkin,
  Zhang, Agarwal, Slama, Ray et~al.}]{r11}
\bibinfo{author}{Ouyang, L.}, \bibinfo{author}{Wu, J.}, \bibinfo{author}{Jiang,
  X.}, \bibinfo{author}{Almeida, D.}, \bibinfo{author}{Wainwright, C.},
  \bibinfo{author}{Mishkin, P.}, \bibinfo{author}{Zhang, C.},
  \bibinfo{author}{Agarwal, S.}, \bibinfo{author}{Slama, K.},
  \bibinfo{author}{Ray, A.} et~al. (\bibinfo{year}{2022}).
\newblock \bibinfo{title}{Training language models to follow instructions with
  human feedback}.
\newblock {\it \bibinfo{journal}{Advances in neural information processing
  systems}\/},  {\it \bibinfo{volume}{35}\/}, \bibinfo{pages}{27730--27744}.
\bibitem[{Park \& Simoff(2014)}]{r45}
\bibinfo{author}{Park, L.~A.}, \& \bibinfo{author}{Simoff, S.}
  (\bibinfo{year}{2014}).
\newblock \bibinfo{title}{Second order probabilistic models for within-document
  novelty detection in academic articles}.
\newblock In {\it \bibinfo{booktitle}{Proceedings of the 37th international ACM
  SIGIR conference on Research \& development in information retrieval}\/} (pp.
  \bibinfo{pages}{1103--1106}).
\bibitem[{Robertson(2023)}]{r19}
\bibinfo{author}{Robertson, Z.} (\bibinfo{year}{2023}).
\newblock \bibinfo{title}{Gpt4 is slightly helpful for peer-review assistance:
  A pilot study}.
\newblock {\it \bibinfo{journal}{arXiv preprint arXiv:2307.05492}\/}, .
\bibitem[{Schulz et~al.(2022)Schulz, Barnett, Bernard, Brown, Byrne, Eckmann,
  Gazda, Kilicoglu, Prager, Salholz-Hillel et~al.}]{r17}
\bibinfo{author}{Schulz, R.}, \bibinfo{author}{Barnett, A.},
  \bibinfo{author}{Bernard, R.}, \bibinfo{author}{Brown, N.~J.},
  \bibinfo{author}{Byrne, J.~A.}, \bibinfo{author}{Eckmann, P.},
  \bibinfo{author}{Gazda, M.~A.}, \bibinfo{author}{Kilicoglu, H.},
  \bibinfo{author}{Prager, E.~M.}, \bibinfo{author}{Salholz-Hillel, M.} et~al.
  (\bibinfo{year}{2022}).
\newblock \bibinfo{title}{Is the future of peer review automated?}
\newblock {\it \bibinfo{journal}{BMC research notes}\/},  {\it
  \bibinfo{volume}{15}\/}, \bibinfo{pages}{203}.
\bibitem[{Schumpeter(1939)}]{r1}
\bibinfo{author}{Schumpeter, J.~A.} (\bibinfo{year}{1939}).
\newblock \bibinfo{title}{Business cycles: a theoretical, historical and
  statistical analysis of the capitalist process}, .
\bibitem[{Shah(2022{\natexlab{a}})}]{r25}
\bibinfo{author}{Shah, N.~B.} (\bibinfo{year}{2022}{\natexlab{a}}).
\newblock \bibinfo{title}{Challenges, experiments, and computational solutions
  in peer review}.
\newblock {\it \bibinfo{journal}{Communications of the ACM}\/},  {\it
  \bibinfo{volume}{65}\/}, \bibinfo{pages}{76--87}.
\bibitem[{Shah(2022{\natexlab{b}})}]{r24}
\bibinfo{author}{Shah, N.~B.} (\bibinfo{year}{2022}{\natexlab{b}}).
\newblock \bibinfo{title}{Improving the peer review process in a scientific
  manner shows promise}.
\newblock {\it \bibinfo{journal}{COMMUNICATIONS OF THE ACM}\/},  {\it
  \bibinfo{volume}{65}\/}, \bibinfo{pages}{75--87}.
\bibitem[{Shen et~al.(2022)Shen, Cheng, Zhou, Bing, You \& Si}]{r39}
\bibinfo{author}{Shen, C.}, \bibinfo{author}{Cheng, L.}, \bibinfo{author}{Zhou,
  R.}, \bibinfo{author}{Bing, L.}, \bibinfo{author}{You, Y.}, \&
  \bibinfo{author}{Si, L.} (\bibinfo{year}{2022}).
\newblock \bibinfo{title}{{MR}e{D}: A meta-review dataset for
  structure-controllable text generation}.
\newblock In \bibinfo{editor}{S.~Muresan}, \bibinfo{editor}{P.~Nakov}, \&
  \bibinfo{editor}{A.~Villavicencio} (Eds.), {\it \bibinfo{booktitle}{Findings
  of the Association for Computational Linguistics: ACL 2022}\/} (pp.
  \bibinfo{pages}{2521--2535}).
\newblock \bibinfo{address}{Dublin, Ireland}: \bibinfo{publisher}{Association
  for Computational Linguistics}.
\newblock \URLprefix \url{https://aclanthology.org/2022.findings-acl.198}.
  \DOIprefix\doi{10.18653/v1/2022.findings-acl.198}.
\bibitem[{Shibayama \& Wang(2020)}]{r68}
\bibinfo{author}{Shibayama, S.}, \& \bibinfo{author}{Wang, J.}
  (\bibinfo{year}{2020}).
\newblock \bibinfo{title}{Measuring originality in science}.
\newblock {\it \bibinfo{journal}{Scientometrics}\/},  {\it
  \bibinfo{volume}{122}\/}, \bibinfo{pages}{409--427}.
\bibitem[{Shibayama et~al.(2021)Shibayama, Yin \& Matsumoto}]{r42}
\bibinfo{author}{Shibayama, S.}, \bibinfo{author}{Yin, D.}, \&
  \bibinfo{author}{Matsumoto, K.} (\bibinfo{year}{2021}).
\newblock \bibinfo{title}{Measuring novelty in science with word embedding}.
\newblock {\it \bibinfo{journal}{PloS one}\/},  {\it \bibinfo{volume}{16}\/},
  \bibinfo{pages}{e0254034}.
\bibitem[{Siler et~al.(2015)Siler, Lee \& Bero}]{r7}
\bibinfo{author}{Siler, K.}, \bibinfo{author}{Lee, K.}, \&
  \bibinfo{author}{Bero, L.} (\bibinfo{year}{2015}).
\newblock \bibinfo{title}{Measuring the effectiveness of scientific
  gatekeeping}.
\newblock {\it \bibinfo{journal}{Proceedings of the National Academy of
  Sciences}\/},  {\it \bibinfo{volume}{112}\/}, \bibinfo{pages}{360--365}.
\bibitem[{Stappen et~al.(2020)Stappen, Rizos, Hasan, Hain \& Schuller}]{r28}
\bibinfo{author}{Stappen, L.}, \bibinfo{author}{Rizos, G.},
  \bibinfo{author}{Hasan, M.}, \bibinfo{author}{Hain, T.}, \&
  \bibinfo{author}{Schuller, B.~W.} (\bibinfo{year}{2020}).
\newblock \bibinfo{title}{Uncertainty-aware machine support for paper reviewing
  on the interspeech 2019 submission corpus}, .
\bibitem[{Sun et~al.(2022)Sun, Barry~Danfa \& Teplitskiy}]{r23}
\bibinfo{author}{Sun, M.}, \bibinfo{author}{Barry~Danfa, J.}, \&
  \bibinfo{author}{Teplitskiy, M.} (\bibinfo{year}{2022}).
\newblock \bibinfo{title}{Does double-blind peer review reduce bias? evidence
  from a top computer science conference}.
\newblock {\it \bibinfo{journal}{Journal of the Association for Information
  Science and Technology}\/},  {\it \bibinfo{volume}{73}\/},
  \bibinfo{pages}{811--819}.
\bibitem[{Tahamtan \& Bornmann(2018)}]{r6}
\bibinfo{author}{Tahamtan, I.}, \& \bibinfo{author}{Bornmann, L.}
  (\bibinfo{year}{2018}).
\newblock \bibinfo{title}{Creativity in science and the link to cited
  references: Is the creative potential of papers reflected in their cited
  references?}
\newblock {\it \bibinfo{journal}{Journal of informetrics}\/},  {\it
  \bibinfo{volume}{12}\/}, \bibinfo{pages}{906--930}.
\bibitem[{Terwiesch(2023)}]{r16}
\bibinfo{author}{Terwiesch, C.} (\bibinfo{year}{2023}).
\newblock \bibinfo{title}{Would chat gpt3 get a wharton mba}.
\newblock {\it \bibinfo{journal}{A prediction based on its performance in the
  operations management course}\/}, .
\bibitem[{Thelwall(2024)}]{ref61}
\bibinfo{author}{Thelwall, M.} (\bibinfo{year}{2024}).
\newblock \bibinfo{title}{Can chatgpt evaluate research quality?}
\newblock {\it \bibinfo{journal}{Journal of Data and Information Science}\/},
  {\it \bibinfo{volume}{9}\/}, \bibinfo{pages}{1--21}. \URLprefix
  \url{https://doi.org/10.2478/jdis-2024-0013}.
  \DOIprefix\doi{doi:10.2478/jdis-2024-0013}.
\bibitem[{Uzzi et~al.(2013)Uzzi, Mukherjee, Stringer \& Jones}]{r5}
\bibinfo{author}{Uzzi, B.}, \bibinfo{author}{Mukherjee, S.},
  \bibinfo{author}{Stringer, M.}, \& \bibinfo{author}{Jones, B.}
  (\bibinfo{year}{2013}).
\newblock \bibinfo{title}{Atypical combinations and scientific impact}.
\newblock {\it \bibinfo{journal}{Science}\/},  {\it \bibinfo{volume}{342}\/},
  \bibinfo{pages}{468--472}.
\bibitem[{Wang et~al.(2023{\natexlab{a}})Wang, Yang, Huang, Yang, Majumder \&
  Wei}]{r59}
\bibinfo{author}{Wang, L.}, \bibinfo{author}{Yang, N.}, \bibinfo{author}{Huang,
  X.}, \bibinfo{author}{Yang, L.}, \bibinfo{author}{Majumder, R.}, \&
  \bibinfo{author}{Wei, F.} (\bibinfo{year}{2023}{\natexlab{a}}).
\newblock \bibinfo{title}{Improving text embeddings with large language
  models}.
\newblock {\it \bibinfo{journal}{arXiv preprint arXiv:2401.00368}\/}, .
\bibitem[{Wang et~al.(2023{\natexlab{b}})Wang, Lin \& Shao}]{r9}
\bibinfo{author}{Wang, X.}, \bibinfo{author}{Lin, X.}, \&
  \bibinfo{author}{Shao, B.} (\bibinfo{year}{2023}{\natexlab{b}}).
\newblock \bibinfo{title}{Artificial intelligence changes the way we work: A
  close look at innovating with chatbots}.
\newblock {\it \bibinfo{journal}{Journal of the Association for Information
  Science and Technology}\/},  {\it \bibinfo{volume}{74}\/},
  \bibinfo{pages}{339--353}.
\bibitem[{Wang et~al.(2024)Wang, Zhang, Chen \& Chen}]{r66}
\bibinfo{author}{Wang, Z.}, \bibinfo{author}{Zhang, H.}, \bibinfo{author}{Chen,
  J.}, \& \bibinfo{author}{Chen, H.} (\bibinfo{year}{2024}).
\newblock \bibinfo{title}{An effective framework for measuring the novelty of
  scientific articles through integrated topic modeling and cloud model}.
\newblock {\it \bibinfo{journal}{Journal of Informetrics}\/},  {\it
  \bibinfo{volume}{18}\/}, \bibinfo{pages}{101587}. \URLprefix
  \url{https://www.sciencedirect.com/science/article/pii/S1751157724000993}.
  \DOIprefix\doi{https://doi.org/10.1016/j.joi.2024.101587}.
\bibitem[{Wei et~al.(2023)Wei, Yuan, Yang, Shen, Li, Wang \& Chen}]{r51}
\bibinfo{author}{Wei, Y.}, \bibinfo{author}{Yuan, S.}, \bibinfo{author}{Yang,
  R.}, \bibinfo{author}{Shen, L.}, \bibinfo{author}{Li, Z.},
  \bibinfo{author}{Wang, L.}, \& \bibinfo{author}{Chen, M.}
  (\bibinfo{year}{2023}).
\newblock \bibinfo{title}{Tackling modality heterogeneity with multi-view
  calibration network for multimodal sentiment detection}.
\newblock In \bibinfo{editor}{A.~Rogers}, \bibinfo{editor}{J.~Boyd-Graber}, \&
  \bibinfo{editor}{N.~Okazaki} (Eds.), {\it \bibinfo{booktitle}{Proceedings of
  the 61st Annual Meeting of the Association for Computational Linguistics
  (Volume 1: Long Papers)}\/} (pp. \bibinfo{pages}{5240--5252}).
\newblock \bibinfo{address}{Toronto, Canada}: \bibinfo{publisher}{Association
  for Computational Linguistics}.
\newblock \URLprefix \url{https://aclanthology.org/2023.acl-long.287}.
  \DOIprefix\doi{10.18653/v1/2023.acl-long.287}.
\bibitem[{Yang et~al.(2019)Yang, Dai, Yang, Carbonell, Salakhutdinov \&
  Le}]{r56}
\bibinfo{author}{Yang, Z.}, \bibinfo{author}{Dai, Z.}, \bibinfo{author}{Yang,
  Y.}, \bibinfo{author}{Carbonell, J.}, \bibinfo{author}{Salakhutdinov, R.~R.},
  \& \bibinfo{author}{Le, Q.~V.} (\bibinfo{year}{2019}).
\newblock \bibinfo{title}{Xlnet: Generalized autoregressive pretraining for
  language understanding}.
\newblock In \bibinfo{editor}{H.~Wallach}, \bibinfo{editor}{H.~Larochelle},
  \bibinfo{editor}{A.~Beygelzimer}, \bibinfo{editor}{F.~d\textquotesingle
  Alch\'{e}-Buc}, \bibinfo{editor}{E.~Fox}, \& \bibinfo{editor}{R.~Garnett}
  (Eds.), {\it \bibinfo{booktitle}{Advances in Neural Information Processing
  Systems}\/}.
\newblock \bibinfo{publisher}{Curran Associates, Inc.}
  volume~\bibinfo{volume}{32}.
\newblock \URLprefix
  \url{https://proceedings.neurips.cc/paper_files/paper/2019/file/dc6a7e655d7e5840e66733e9ee67cc69-Paper.pdf}.
\bibitem[{Yin et~al.(2023)Yin, Wu, Yokota, Matsumoto \& Shibayama}]{r44}
\bibinfo{author}{Yin, D.}, \bibinfo{author}{Wu, Z.}, \bibinfo{author}{Yokota,
  K.}, \bibinfo{author}{Matsumoto, K.}, \& \bibinfo{author}{Shibayama, S.}
  (\bibinfo{year}{2023}).
\newblock \bibinfo{title}{Identify novel elements of knowledge with word
  embedding}.
\newblock {\it \bibinfo{journal}{Plos one}\/},  {\it \bibinfo{volume}{18}\/},
  \bibinfo{pages}{e0284567}.
\bibitem[{Yuan et~al.(2022)Yuan, Liu \& Neubig}]{r36}
\bibinfo{author}{Yuan, W.}, \bibinfo{author}{Liu, P.}, \&
  \bibinfo{author}{Neubig, G.} (\bibinfo{year}{2022}).
\newblock \bibinfo{title}{Can we automate scientific reviewing?}
\newblock {\it \bibinfo{journal}{Journal of Artificial Intelligence
  Research}\/},  {\it \bibinfo{volume}{75}\/}, \bibinfo{pages}{171--212}.
\bibitem[{Yue et~al.(2019)Yue, Chen, Li, Zuo \& Yin}]{r71}
\bibinfo{author}{Yue, L.}, \bibinfo{author}{Chen, W.}, \bibinfo{author}{Li,
  X.}, \bibinfo{author}{Zuo, W.}, \& \bibinfo{author}{Yin, M.}
  (\bibinfo{year}{2019}).
\newblock \bibinfo{title}{A survey of sentiment analysis in social media}.
\newblock {\it \bibinfo{journal}{Knowledge and Information Systems}\/},  {\it
  \bibinfo{volume}{60}\/}, \bibinfo{pages}{617--663}.
\bibitem[{Zhao \& Zhang(2025)}]{r72}
\bibinfo{author}{Zhao, Y.}, \& \bibinfo{author}{Zhang, C.}
  (\bibinfo{year}{2025}).
\newblock \bibinfo{title}{A review on the novelty measurements of academic
  papers}.
\newblock {\it \bibinfo{journal}{Scientometrics}\/},  {\it
  \bibinfo{volume}{130}\/}, \bibinfo{pages}{727--753}. \URLprefix
  \url{https://doi.org/10.1007/s11192-025-05234-0}.
  \DOIprefix\doi{10.1007/s11192-025-05234-0}.

\end{thebibliography}
\section{Appendix}
\label{appendix}
\renewcommand*{\thefigure}{S\arabic{figure}}
\subsection{Feedback for other scores}\label{app1}
\begin{figure*}[p]%
	\centering
	\includegraphics[width=1\textwidth]{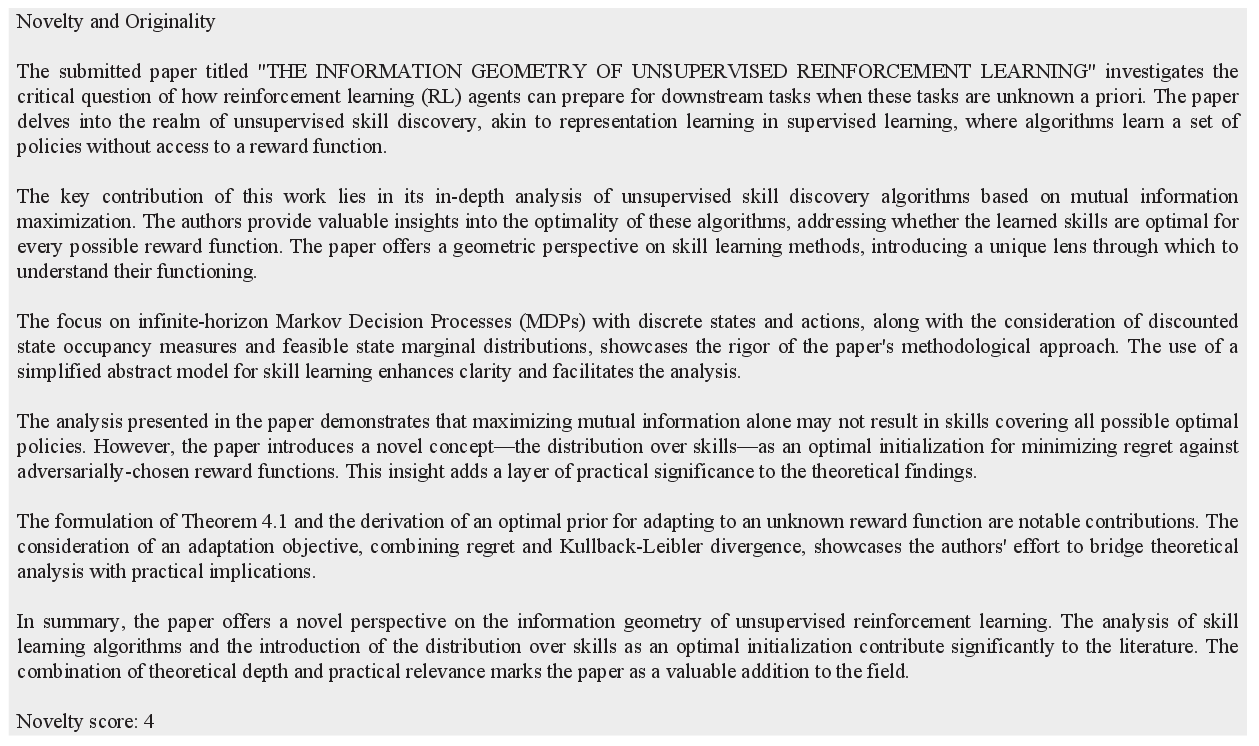}
	\vspace{-0.9cm}
	\caption{\centering{ChatGPT provides feedback based on the prompt in Figure \ref{fig:1}. Novelty score from review report is 4.}}
	\label{fig:s1}
\end{figure*}

\begin{figure*}[p]%
	\centering
	\includegraphics[width=1\textwidth]{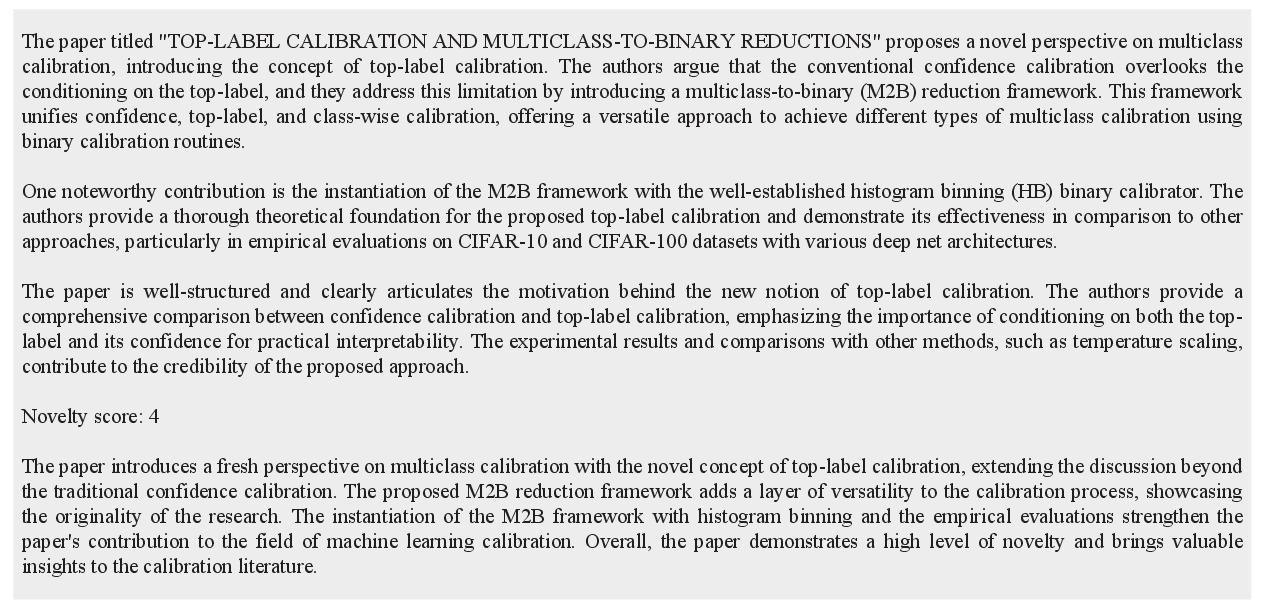}
	\vspace{-0.9cm}
	\caption{\centering{ChatGPT provides feedback based on the prompt in Figure \ref{fig:1}. Novelty score from review report is 2.}}
	\label{fig:s2}
\end{figure*}

\begin{figure*}[t]%
	\centering
	\includegraphics[width=1\textwidth]{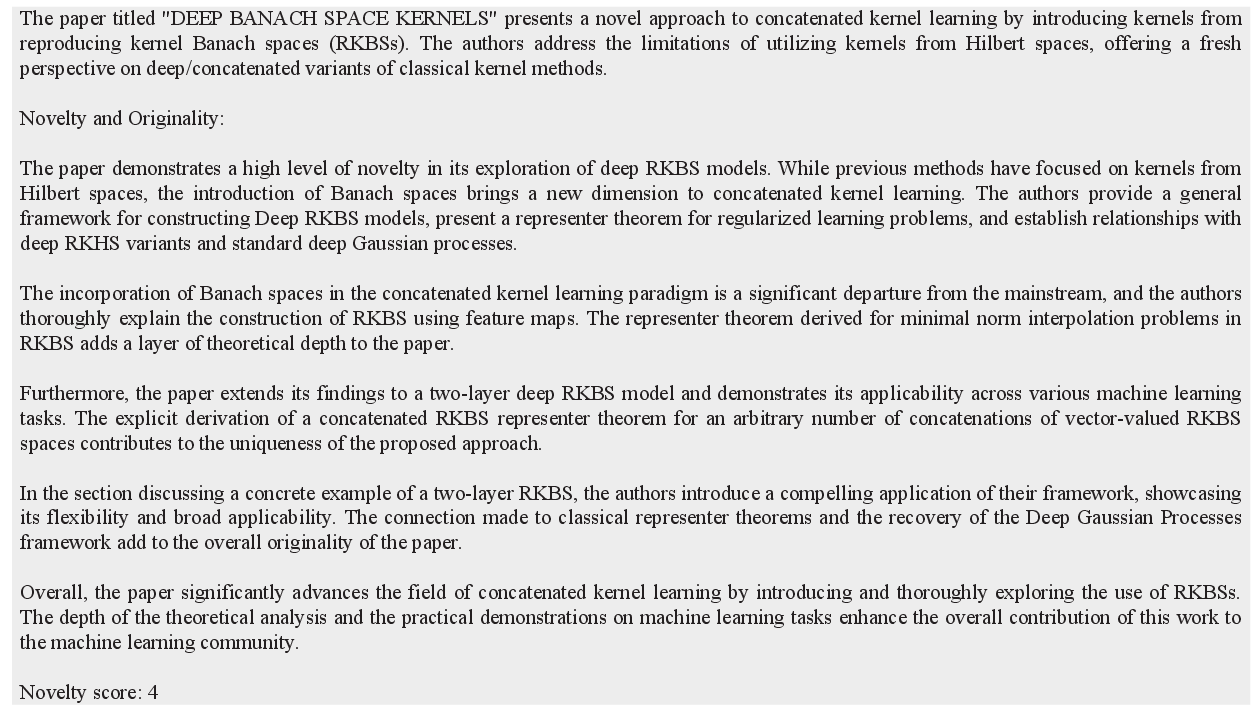}
	\vspace{-0.9cm}
	\caption{\centering{ChatGPT provides feedback based on the prompt in Figure \ref{fig:1}. Novelty score from review report is 1.}}
	\label{fig:s3}
\end{figure*}

The following figures depict the feedback provided by ChatGPT for different novelty scores. As shown in Figure \ref{fig:s1}, \ref{fig:s2} and \ref{fig:s3}.


\subsection{TNS Definition}\label{app2}
Table \ref{tabs1} provides definitions for each score in Technical Novelty and Significance score (TNS).
\begin{table}[h]
	\caption{\centering{Definition of Technical Novelty and Significance score.}}\label{tabs1}%
	\centering
	\vspace{0.2cm}
	\begin{tabular}{@{}m{30mm}<{\centering} m{80mm}<{\centering} @{}}
		\toprule
		\textbf{TNS} & \textbf{Definition}  \\
		\midrule
		\# TNS=1    & The contributions are neither significant nor novel.\\
		\# TNS=2    & The contributions are only marginally significant or novel.  \\
		\# TNS=3    & The contributions are significant and somewhat new. Aspects of the contributions exist in prior work. \\
		\# TNS=4    & The contributions are significant, and do not exist in prior works. \\
		\bottomrule 
	\end{tabular}
\end{table}
\subsection{Prompt of large language models baseline}\label{app3}
\begin{figure*}[t]%
	\centering
	\includegraphics[width=1\textwidth]{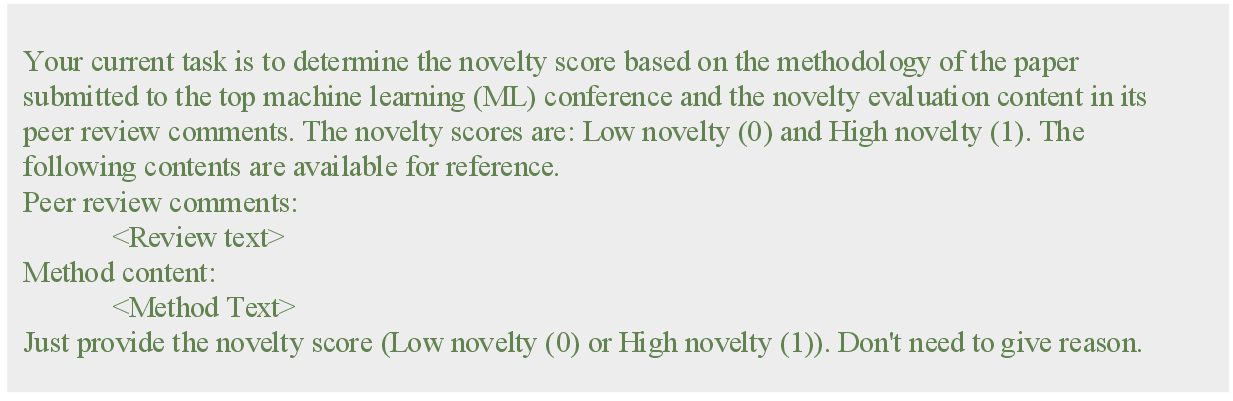}
	\vspace{-0.7cm}
	\caption{\centering{Prompt template of LLMs for review text and method text.}}
	\label{fig:s4}
\end{figure*}
\begin{figure*}[t]%
	\centering
	\includegraphics[width=1\textwidth]{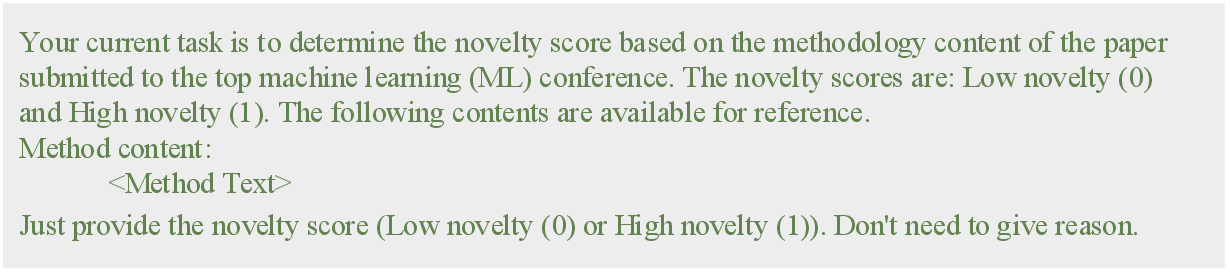}
	\vspace{-0.7cm}
	\caption{\centering{Prompt template of LLMs for method text.}}
	\label{fig:s5}
\end{figure*}
\begin{figure*}[t]%
	\centering
	\includegraphics[width=1\textwidth]{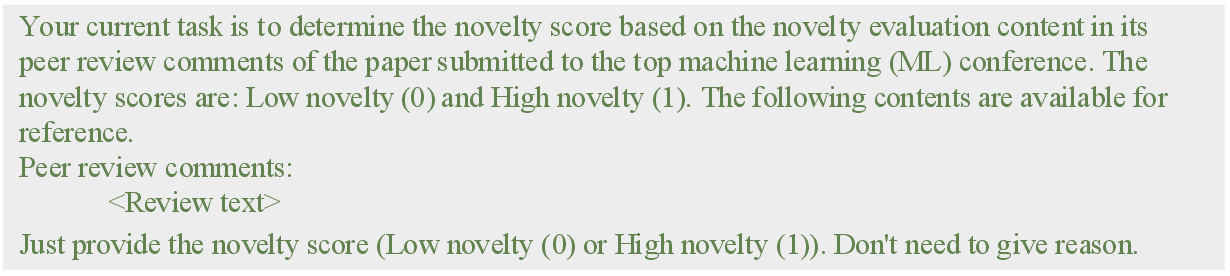}
	\vspace{-0.7cm}
	\caption{\centering{Prompt template of LLMs for review text.}}
	\label{fig:s6}
\end{figure*}
The prompt templates for LLMs are shown in Figure \ref{fig:s4}, \ref{fig:s5} and \ref{fig:s6}. The results generated by the LLMs in this paper can be found at \href{https://github.com/njust-winchy/method_novelty_predict/blob/main/LLM_results}{here}.

\end{document}